\documentclass[10pt,journal,compsoc]{IEEEtran}
\usepackage{graphicx}
\usepackage{booktabs}
\usepackage{subcaption}
\usepackage{epstopdf}
\usepackage{diagbox}
\usepackage{threeparttable}
\usepackage{makecell,multirow,diagbox}

\usepackage[linesnumbered,ruled,vlined]{algorithm2e}
\usepackage{algpseudocode}
\usepackage{amssymb}
\usepackage{amsmath}
\usepackage{amsthm}
\usepackage{cite}
\usepackage{hyperref}
\usepackage{bm}
\usepackage{mathtools}
\usepackage{pgfmath}
\usepackage{dsfont}
\usepackage{xcolor}
\usepackage{color}
\usepackage{colortbl}
\usepackage{bbding}
\usepackage{flushend}
\usepackage{pifont}

\ifCLASSINFOpdf

\else

\fi

\begin{document}

\title{NestQuant: Post-Training Integer-Nesting Quantization for On-Device DNN}

\author{Jianhang~Xie,
        ~Chuntao~Ding,
        ~Xiaqing~Li,
        ~Shenyuan~Ren,
        ~Yidong~Li,
        ~Zhichao~Lu
        \\
\IEEEcompsocitemizethanks{
\IEEEcompsocthanksitem Jianhang Xie, Xiaqing Li, Shenyuan Ren and Yidong Li are with the Key Laboratory of Big Data \& Artificial Intelligence in Transportation (Beijing Jiaotong University), Ministry of Education, with the School of Computer Science and Technology, Beijing Jiaotong University, Beijing, China.
E-mail: \{xiejianhang, xqli1, syren, ydli\}@bjtu.edu.cn.

\IEEEcompsocthanksitem Chuntao Ding is with the School of Artificial Intelligence, Beijing Normal University, Beijing, China.
E-mail: ctding@bnu.edu.cn

\IEEEcompsocthanksitem Zhichao Lu is with the Department of Computer Science, City University of Hong Kong, HKSAR.
                E-mail: luzhichaocn@gmail.com.\\
}
}

\markboth{IEEE Transactions on Mobile Computing}%
{Shell \MakeLowercase{\textit{et al.}}: Bare Demo of IEEEtran.cls for IEEE Transactions on Magnetics Journals}


\IEEEtitleabstractindextext{%
\begin{abstract}
Deploying quantized deep neural network (DNN) models with resource adaptation capabilities on ubiquitous Internet of Things (IoT) devices to provide high-quality AI services can leverage the benefits of compression and meet multi-scenario resource requirements.
However, existing dynamic/mixed precision quantization requires retraining or special hardware, whereas post-training quantization (PTQ) has two limitations for resource adaptation:
(i) The state-of-the-art PTQ methods only provide one fixed bitwidth model, which makes it challenging to adapt to the dynamic resources of IoT devices;
(ii) Deploying multiple PTQ models with diverse bitwidths consumes large storage resources and switching overheads.
To this end, this paper introduces a resource-friendly post-training integer-nesting quantization, i.e., NestQuant, for on-device quantized model switching on IoT devices.
The proposed NestQuant incorporates the integer weight decomposition, which bit-wise splits quantized weights into higher-bit and lower-bit weights of integer data types. 
It also contains a decomposed weights nesting mechanism to optimize the higher-bit weights by adaptive rounding and nest them into the original quantized weights.
In deployment, we can send and store only one NestQuant model and switch between the full-bit/part-bit model by paging in/out lower-bit weights to adapt to resource changes and reduce consumption.
Experimental results on the ImageNet-1K pretrained DNNs demonstrated that the NestQuant model can achieve high performance in top-1 accuracy, and reduce in terms of data transmission, storage consumption, and switching overheads.
In particular, the ResNet-101 with INT8 nesting INT6 can achieve 78.1\% and 77.9\% accuracy for full-bit and part-bit models, respectively, and reduce switching overheads by approximately 78.1\% compared with diverse bitwidths PTQ models.
Code: \url{https://github.com/jianhayes/NESTQUANT}.

 \end{abstract}
}

\maketitle

%
\IEEEpeerreviewmaketitle

\section{Introduction}
\noindent
\IEEEPARstart{O}n-device deploying deep neural network (DNN) models has become mainstream providing AI-powered deep learning (DL) services~\cite{xu2020deepwear, Liu@ANew, xu2019Afirst, zhang2024Acomprehensive, zhang2024secaas, huang2024multi, yao2024intelligent} on ubiquitous Internet of Things (IoT) devices, such as smartphones, wearables, and advanced driving systems, etc.
Nevertheless, high-performance DNNs always consume a huge amount of memory and computational resources, which often makes deploying them on resource-constrained IoT devices infeasible.
Fortunately, \emph{model quantization} technologies~\cite{Han@Deep, Hubara@Quantized, banner2019post, jin2020adabits, yu2021any, xu2023eq, liu24spark, nagel2020up, hubara2021accurate, li2021brecq, frantar2022optimal, guo2022squant} can streamline the models by transforming the floating-point (FP) data type parameters into integer (INT) data type parameters for compression.
Deploying quantized models to solve the storage or communication resource constraints of IoT devices has attracted significant interest from both industry and academia~\cite{Stefanos@SPINN, yang2022cnnpc, chen2023energy, zhang23lightfr, chen2024comm}.
\begin {figure}[t]
\centering
\includegraphics[width=0.9\linewidth]{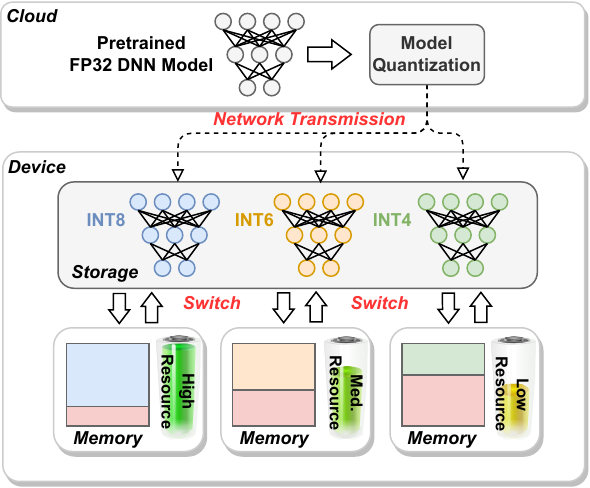}
\caption {Existing system architecture for on-device deployment of diverse bitwidths models for resource adaptation.} 
\label{fig:problem}
\vspace{-0.5cm}
\end{figure}

\vspace{3pt}
\noindent\textbf{Motivation.} Dynamic resource adaptation is important for on-device AI applications, as the availability of resources on IoT devices can vary over time (such as power and storage capacity).
For example, when a smartphone's battery is sufficiently charged, we may opt to run applications in a high-performance mode with a smooth user experience.
As the battery level diminishes to a certain threshold (e.g., 50\%), to prolong the duration, switching the model to an energy-saving mode can conserve power, albeit at the cost of reduced performance.
Dynamic resource adaptation via the perspective of model quantization can further adapt to multiple scenarios with the advantages of compression.

\vspace{3pt}
\noindent\textbf{Challenges.} However, existing dynamic precision quantization methods~\cite{jin2020adabits, yu2021any, xu2023eq} require highly resource-intensive retraining, and 
the mixed precision quantization method~\cite{liu24spark} needs special hardware.
For the post-training quantization:
(i) The state-of-the-art post-training quantization methods for DNNs~\cite{nagel2020up, hubara2021accurate, li2021brecq, frantar2022optimal, guo2022squant} only provide one fixed bitwidth model, which limits the model to adapt to multi-scenarios.
(ii) Storing diverse bitwidth models on IoT devices allows switching models to adapt to different resource requirements, as depicted in Fig.~\ref{fig:problem}. 
However, this approach consumes more storage space and increases memory page-in/-out overheads. 
Additionally, transmitting many models will burden the cloud and device network bandwidth.

\vspace{3pt}
\noindent\textbf{Our Solution.} 
To tackle aforementioned challenges, we propose a Post-Training Integer-\textbf{Nest}ing \textbf{Quant}ization (NestQuant): a quantized DNN model switching mechanism, which leverages the similarity between higher-bit and original weights to switch between full-bit/part-bit models via paging in/paging out lower-bit weights to adapt to IoT devices' dynamic resources changing.

First, we are the first to observe that the higher bits in integer parameters of quantized models exhibit significant similarity of original weights. Specifically, the higher bits can be extracted to form a new bitwidth model while retaining partially usable performance.
Based on these preconditions, we propose an \emph{Integer Weight Decomposition} for integer data type weights, which splits the original weights into two parts: the higher- and lower-bit weights.
We demonstrate that higher-bit weights can also be optimized by Hessian-based adaptive rounding for increasing performance.

Then, we propose the \emph{NestQuant}, which contains (i) a decomposed weights nesting mechanism of higher- and lower-bit weights recomposition and (ii) model switching between full-bit/part-bit models.
The NestQuant model can only store two decomposed weights instead of the original weights.
It can switch to the full-bit model for high-performance inference when memory resources are adequate.
When memory resources are temporarily constrained, the full-bit model can downgrade to the part-bit model by paging out lower-bit weights to ensure the balance between service continuity and performance.
To upgrade to the full-bit model, it can page in the lower-bit weights and recompose them with higher-bit weights.

Finally, to verify our proposed methods, we conduct experiments on the ImageNet-1K pretrained DNNs for evaluating the NestQuant model performance, including the convolutional neural network (CNN) and Vision Transformer (ViT) structures.
The experimental results show that the NestQuant not only retains full-bit model performance but also achieves usable accuracy in part-bit model on ImageNet-1K image classification.
We deploy the NestQuant model with packed-bit tensor technology for different bitwidths.
The on-device resource measurements of model size, network traffic, and numerically calculated switching overheads also demonstrate the resource efficiency of the NestQuant model.

In summary, our main contributions are as follows:
\begin{itemize}
\item This paper proposes an integer weight decomposition for bit-wise splitting the quantized weights into two decomposed weights.
We identify the similarity between higher-bit and original weights and demonstrate that higher-bit weights can be optimized by adaptive rounding.
Thus, the higher-bit weights can be extracted to form a new quantized model.

\item This paper proposes NestQuant for on-device quantized model switching.
The part-bit model is nested into the full-bit model, so we can only store decomposed weights.
The full-bit model can downgrade to part-bit one by paging out lower-bit weights when the resource is insufficient.
The part-bit model can upgrade to the full-bit model with adequate resources by paging in lower-bit weights.

\item This paper conducts experiments on ImageNet pretrained DNNs' performance evaluations and on-device resource measurements. 
The results demonstrate that the proposed NestQuant model outperforms in terms of accuracy, model size, network traffic in transmission, and switching overheads.

\end{itemize}

The rest of the paper is organized as follows. 
First, Section~\ref{ref:2-related} reviews the relevant literature. 
Section~\ref{ref:3-approach} describes the NestQuant in detail. 
Section~\ref{ref:4-experiment} presents the experimental results. 
Finally, we conclude our paper in Section~\ref{ref:5-conclusion}.

\section{Background and Related Work} \label{ref:2-related} 

\subsection{Model Quantization} \label{ref:2-quant} 
To achieve DNN model compression for efficient on-device DL, model quantization is more popular than 
other compression technologies because of the generalizability of integer data types for different hardware.

Model quantization~\cite{Han@Deep, Hubara@Quantized, banner2019post} is to approximate the FP tensor parameters $\bm{x}$ as the multiplication of INT tensor parameters $\bm{x}_\mathrm{int}$ and scales $s$, i.e., $\bm{x}\!\approx\!\hat{\bm{x}}\!=\!s \cdot \bm{x}_\mathrm{int}$, the tensor parameters can be the weights or activations in model.
The storage and memory page-in/-out overheads can be saved because the FP32 model is compressed into the INT model.
Quantization algorithms can be categorized as the \emph{Quantization-Aware Training} (QAT) and \emph{Post-Training Quantization} (PTQ).
The former requires re-training the model, and the latter does not and only needs some calibration data or data-free to optimize the tensor parameters. 
In addition, \emph{Dynamic/Mixed Precision Quantization}\cite{jin2020adabits, yu2021any, xu2023eq, liu24spark} requires retraining or special hardware.
As a result, PTQ methods are preferred for on-device DNN compression and deployment.

\vspace{3pt}
\noindent{\bf{Hessian-based Post-Training Quantization:}}
The current state-of-the-art PTQ is the Hessian-based optimization, 
where the difference of task loss $\delta\mathcal{L}\!=\!\mathcal{L}(\bm{w})\!-\!\mathcal{L}(\hat{\bm{w}})$ is assumed to be a function of the weight quantization perturbations $\delta \bm{w}\!=\!(\bm{w}\!-\!\hat{\bm{w}})/s$, and $\hat{\bm{w}}\!=\!s \cdot \bm{w}_\mathrm{int}$.
The $\delta\mathcal{L}(\delta \bm{w})$ can have a second-order Taylor series approximation, for the well-trained converged DNN model, the first-order item of gradients is $\mathbf{g}\!=\!\bf{0}$, and the second-order item is the Hessian matrix of weight $\bm{w}$, as shown in Equation~\ref{eq:hessian_approx}:
{
\begin{equation}
\begin{aligned}
\mathrm{\bf{\mathbb{E}}}\left[\delta\mathcal{L}(\delta \bm{w})\right]&\approx \mathrm{\bf{\mathbb{E}}}\left[\mathbf{g}^\mathrm{T}\delta \bm{w} + \frac{1}{2}\delta \bm{w}^\mathrm{T} \mathbf{H}^{(\bm{w})} \delta \bm{w}\right]\\
&\approx\mathrm{\bf{\mathbb{E}}}\left[\frac{1}{2}\delta \bm{w}^\mathrm{T} \mathbf{H}^{(\bm{w})} \delta \bm{w}\right]
\end{aligned}.
\label{eq:hessian_approx}
\end{equation}
}%
\noindent There are many PTQ algorithms~\cite{nagel2020up, hubara2021accurate, li2021brecq, frantar2022optimal, guo2022squant} work to approximate and optimize the weight Hessian $\mathbf{H}^{(\bm{w})}$ to minimize the $\delta\mathcal{L}$ to retain the model performance.
For example, Nagel \emph{et al.}~\cite{nagel2020up} proposed \emph{AdaRound} to approximate the weight Hessian as layer-wise constant diagonal matrices.
Li \emph{et al.}~\cite{li2021brecq} introduced \emph{BRECQ} with the Fisher Information to approximate the Hessian and block-wise optimization.
Frantar \emph{et al.}~\cite{frantar2022optimal} proposed \emph{OBQ} with an iterative Hessian approximation for PTQ optimization.
Guo \emph{et al.}~\cite{guo2022squant} proposed \emph{SQuant} with a Hessian decomposition approximation to achieve a data-free weights adaptive rounding.
These Hessian-based PTQ algorithms have excellent performance but with computationally intensive optimization requirements of Hessian approximation, and are hard to process on resource-constrained IoT devices.

\subsection{Resource-Aware On-Device Model}

There are many resource-aware on-device models ~\cite{fang2018nestdnn, Fang@FlexDNN, wen2023adaptivenet, ding2021resource, ding2022resource, ding2022edge}.
These models are about the neural architecture switching for resource adaptation.
However, these methods all require labor-intensive training from scratch for constructing a switchable DNN supernet, which will also consume a large disk space to store.
Inspired by these works, we consider achieving the resource adaptation from completely different perspectives, i.e., \emph{Nesting Model Quantization} and \emph{Post-Training}.
We consider introducing the nesting idea into the PTQ to get an on-device quantized model switching for achieving an easy application of resource adaptation without training and redesigning.
However, the neural architecture or feature-aspect nesting is tolerant of performance degradation, and the integer-aspect nesting is changing the numeric structure of parameters.
The changes in numerical values are more sensitive to model performance, which is a challenge in the design of nesting integer data types.

\subsection{Computation and On-Device DL Library Limitation} \label{ref:2-limit} 
\noindent{\bf{IoT Device Computation Limitation:}}
Existing PTQ algorithms are still computationally intensive as our evaluation of 8-bit weight and activation PTQ, as depicted in Table~\ref{tab:speed}.
\begin{table}[ht]
\centering
\caption{ResNet-18 W8A8 PTQ Evaluation.\label{tab:speed}}
\resizebox{.45\textwidth}{!}{%
\begin{tabular}{lcccc}
\toprule
\multirow{2}*{\makecell[c]{Hardware}}& 
\multirow{2}*{\makecell[c]{PTQ Algorithm}}& 
\multirow{2}*{\makecell[c]{Optim. Time}}&
\multirow{2}*{\makecell[c]{Max GPU\\Mem. Usage}}&
\multirow{2}*{\makecell[c]{Require\\Data}}\\
\\
\midrule
\multirow{4}*{\makecell[c]{RTX 2080Ti\\Edge Server}}
& BRECQ & 1901 Sec & 4.3GB & \CheckmarkBold \\
& OBQ & 5187 Sec & 7.8GB & \CheckmarkBold \\
& SQuant (parallel) & 2 Sec & 6.8GB & \XSolidBrush \\
& SQuant (serial) & 241 Sec & 6.8GB & \XSolidBrush \\
\midrule
\multirow{1}*{\makecell[c]{Raspberry Pi 4B}}& 
\multirow{1}*{\makecell[c]{SQuant (serial)}} &
\multirow{1}*{\makecell[c]{1445 Sec}} &
\multirow{1}*{\makecell[c]{-}} & 
\multirow{1}*{\makecell[c]{\XSolidBrush}} \\
\bottomrule
\end{tabular}%
}
\end{table}

The IoT devices do not have such abundant resources to support the optimization as shown in Table~\ref{tab:hardware}. 
For example, Raspberry Pi 3B+, 4B, and Jetson Nano B01 have computation performance~\cite{machinesGFLOPS, jetsonhardware} of 5.3, 9.69, and 472 GFLOPS, respectively, which are approximately 2500$\times$, 1400$\times$, and 30$\times$ lower than the server with RTX 2080Ti GPU.
PTQ optimizations are hard to process, and the re-training is even less feasible, so these devices are only suitable for inference. 
\begin{table}[ht]
\centering
\caption{Common Hardware Resource Conditions.\label{tab:hardware}}
\resizebox{.45\textwidth}{!}{%
\begin{tabular}{lccccc}
\toprule
\multirow{3}*{\makecell[c]{Hardware}}&
\multirow{3}*{\makecell[c]{RTX 2080Ti\\Edge Server}} & \multicolumn{3}{c}{IoT Devices}\\ \cmidrule{3-5}
&&
\multirow{2}*{\makecell[c]{Jetson\\Nano B01}}& 
\multirow{2}*{\makecell[c]{Raspberry\\Pi 4B}} & 
\multirow{2}*{\makecell[c]{Raspberry\\Pi 3B+}}\\
\\
\midrule
\multirow{2}*{\makecell[c]{Comput. Perf.}}& 
\multirow{2}*{\makecell[c]{13.4\\TFLOPS}} &
\multirow{2}*{\makecell[c]{472\\GFLOPS}} & 
\multirow{2}*{\makecell[c]{9.69\\GFLOPS}} & 
\multirow{2}*{\makecell[c]{5.3\\GFLOPS}}\\
\\
\midrule
\multirow{1}*{\makecell[c]{Mem./GPU Mem.}}&
\multirow{1}*{\makecell[c]{64GB/11GB}} &
\multirow{1}*{\makecell[c]{4GB}} & 
\multirow{1}*{\makecell[c]{4GB}} & 
\multirow{1}*{\makecell[c]{4GB}}\\
\bottomrule
\end{tabular}%
}
\end{table}

The resources of IoT devices are commonly changing dynamically, e.g., memory page-in/-out, transmission bandwidth, and storage.
With the on-device deployment of diverse bitwidths models, the server needs repeated quantizing and sending models, the device needs to store multiple models, which costs a huge amount of memory page-in/-out overheads for model switching.
%

\vspace{3pt}
\noindent{\bf{On-Device DL Library Limitation:}}
For representative on-device DL libraries, such as \emph{TFLite}, \emph{PyTorchMobile}, and \emph{Ncnn}~\cite{tflite,pytorchmobile,ncnn} do not support the lower integer data type currently, i.e., lower than 8-bit, even \emph{ONNX} and \emph{ONNX Runtime}~\cite{onnx, onnxruntime} just support 4-bit data types with 2$\times$4-bit packing only in recent versions, as shown in Table~\ref{tab:dl_libraries}.
\begin{table}[ht]
\centering
\caption{Representative DL libraries Data Type Supports.\label{tab:dl_libraries}}
\resizebox{.45\textwidth}{!}{%
\begin{tabular}{lc}
\toprule
\multirow{1}*{\makecell[l]{Libraries}} 
&\multirow{1}*{\makecell[l]{Quantized Data Type}} \\
\midrule
TensorFlow/TFLite & quint32, quint16, qint16, quint8, qint8 \\
PyTorch/PyTorchMobile & quint8, qint8, quint4x2 \\
ONNX/ONNX Runtime & uint8, int8, uint4x2, int4x2 \\  
Ncnn & int8 \\ \bottomrule
\end{tabular}%
}
\end{table}

Therefore, there is a gap in the on-device applications of quantized models with lower bitwidths.
However, our work has great potential for the future if the DL libraries support lower data types (e.g., 1-bit to 7-bit).
As a result, we will verify the feasibility of the model size with NestQuant by arbitrary-bit data type weights in the form of the packed-bits tensors~\cite{petersen2022difflogic, petersen2023distributional} for compromise.

\section{The proposed Method: NestQuant} \label{ref:3-approach} 
In this section, we first introduce the preliminaries of quantization optimization.
Then, we propose the integer weight decomposition, the theoretical foundation of NestQuant, which demonstrates that the decomposed higher-bit weights can be extracted to form a new quantized model.
For this, we propose the NestQuant, which can switch between the full-bit/part-bit model by storing decomposed weights.
For the part-bit model, we analyze the effective nested combinations. 
For the full-bit model, we introduce an extra 1-bit to compensate for the performance.
Finally, we discuss the practical selection and deployment of the NestQuant model.

We describe the proposed NestQuant in detail as follows: (i) preliminaries in Section~\ref{ref:quant_optim_prereq}, (ii) theoretical foundation of NestQuant in Section~\ref{ref:info_anaysis}, and (iii) procedures, switching instances, and implementation of NestQuant in Section~\ref{ref:nestquant_framework}.

\subsection{Preliminaries} \label{ref:quant_optim_prereq}
The simplest linear quantization \emph{min-max} computes the scale by the minimum and maximum of weights, and quantizes FP32 weights and activations to the INT data type by rounding-to-nearest (RTN).
In this paper, we discuss symmetric linear quantization with signed INT weights and ignore the zero-point.
We first specifically use the $\bm{w}$ to denote the weight variables, the scale of quantized parameters is $s$, and the INT data type weight is $\bm{w}_\mathrm{int}$.
The quantization of the INT$n$ weight $\bm{w}$ can be described as:
{
\begin{equation}
\begin{aligned}
\bm{w}_\mathrm{int}\!=\!\mathrm{Clip}\left(\left\lfloor\frac{\bm{w}}{s}\right\rceil, min, max\right)
\end{aligned},
\label{eq:quantization}
\end{equation}
}%
\noindent where the $\mathrm{Clip}\left(\ \cdot \ , min, max\right)$ is the clipping function, the upper and lower thresholds of clipping are $min=-2^{n-1}$ and $max=2^{n-1}-1$, which are defined by the types of signed INT number, the $\lfloor\cdot\rceil$ is the rounding operator, and dequantization of the weight $\bm{w}_\mathrm{int}$ to the FP32 weight can be described as:
{
\begin{equation}
\begin{aligned}
\hat{\bm{w}}\!=\!s\cdot \bm{w}_\mathrm{int}
\end{aligned}.
\label{eq:dequantization}
\end{equation}
}%
\noindent The weight quantization perturbations can be defined as:
{
\begin{equation}
\begin{aligned}
\delta \bm{w}\!=\frac{\!\bm{w}\!-\!\hat{\bm{w}}}{s}\!=\!\frac{\bm{w}}{s}\!-\!\bm{w}_\mathrm{int}
\end{aligned},
\label{eq:perturbation}
\end{equation}
}%
\noindent and the popular way to optimize the quantization tensor is minimizing the mean squared error or Kullback-Leibler Divergence between the original weight $\bm{w}$ and the dequantized one $\hat{\bm{w}}$ by re-training or fine-tuning.

Considering the high cost of re-training and data privacy protection of original training datasets, we focus on the PTQ algorithms. 
The state-of-the-art PTQ algorithms are Hessian-based PTQ~\cite{nagel2020up, hubara2021accurate, li2021brecq, guo2022squant, frantar2022optimal}.
It means minimizing the approximation of Hessian $\mathbf{H}^{(\bm{w})}$ as a layer-wise Hessian matrix $\mathbf{H}^{(\bm{w}^{(\ell)})}$ of the weight perturbations $\delta \bm{w}$ for layer $\ell$ in Equation~\ref{eq:hessian}, which is equivalently minimizing the task loss degradation due to the quantization perturbation of weight, as described in Equation~\ref{eq:hessian_approx}. 
{
\begin{equation}
\begin{aligned}
\mathop{\mathrm{argmin}}\limits_{\delta \bm{w}^{(\ell)}}\mathrm{\bf{\mathbb{E}}}\left[ \delta \bm{w}^{(\ell)^\mathrm{T}} \mathbf{H}^{(\bm{w}^{(\ell)})} \delta \bm{w}^{(\ell)} \right]
\end{aligned}.
\label{eq:hessian}
\end{equation}
}

However, the Hessian-based optimization is still a computationally intensive burden, which is hard to process on IoT devices, as discussed in Section~\ref{ref:2-limit}.
Meanwhile, PTQ algorithms can only provide one fixed bitwidth model, which makes it impossible to provide smooth and continuous on-device services on changing resources.

Optimizing and storing diverse bitwidths PTQ models (e.g., storing an INT8 and an INT4 model is referred to as INT8+INT4) would take up an amount of storage space, as well as the extra overheads of switching on the device.
As a result, it is a necessity to design a more efficient PTQ model switching mechanism for resource adaptation.
\begin {figure}[t]
\centering
\includegraphics[width=0.75\linewidth]{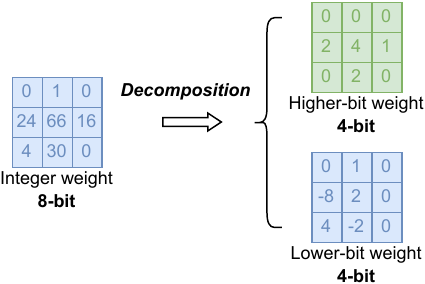}
\caption {Illustration of integer weight decomposition.}
\label{fig:iwd}
\end{figure}

\begin{figure*}[ht]
    \begin{subfigure}[b]{0.25\textwidth}
    \centering
    \includegraphics[trim={0, 0, 0, 0}, clip, width=\textwidth]{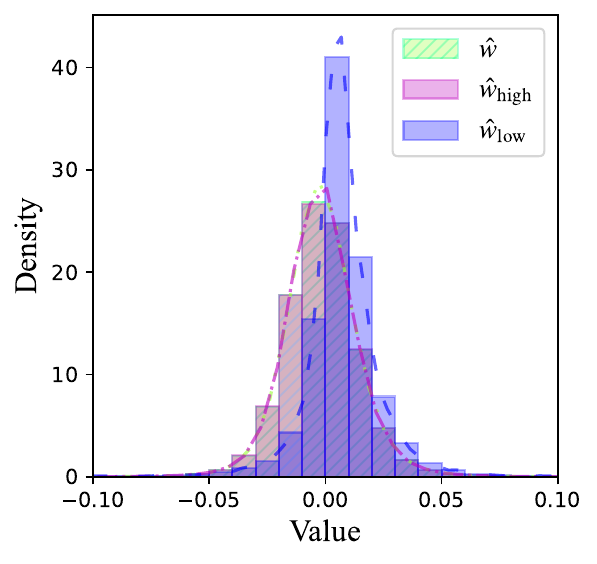}
    \caption{INT(8\textbar5)\label{fig:kde_int8_5}}
    \end{subfigure}\hfill
    \begin{subfigure}[b]{0.25\textwidth}
    \centering
    \includegraphics[trim={0, 0, 0, 0}, clip, width=\textwidth]{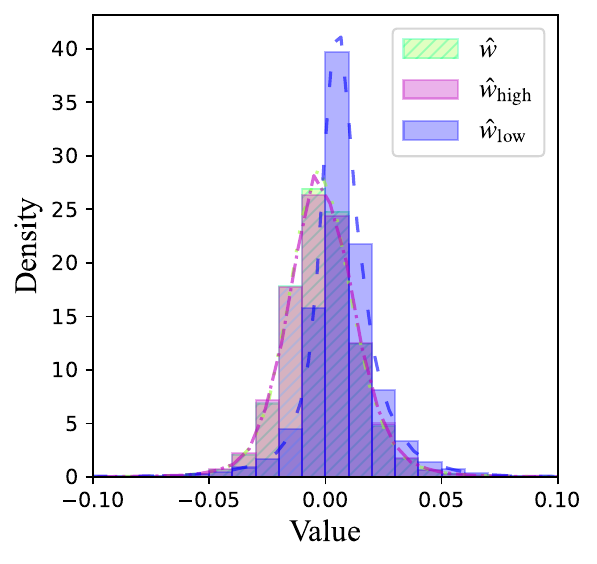}
    \caption{INT(8\textbar4)\label{fig:kde_int8_4}}
    \end{subfigure}\hfill
    \begin{subfigure}[b]{0.25\textwidth}
    \centering
    \includegraphics[trim={0, 0, 0, 0}, clip, width=\textwidth]{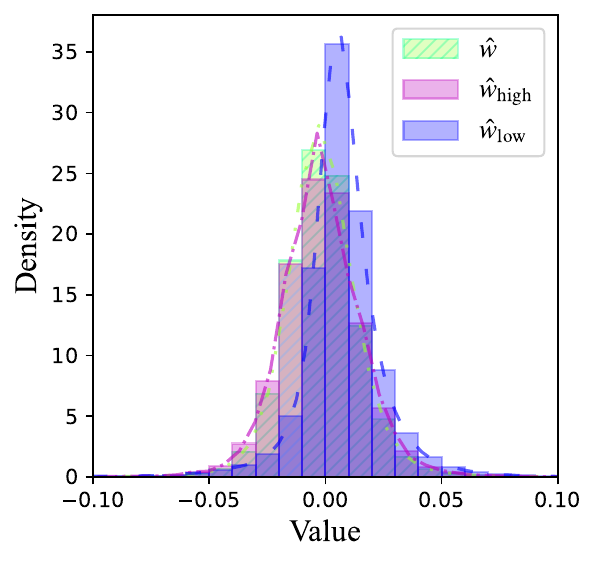}
    \caption{INT(8\textbar3)\label{fig:kde_int8_3}}
    \end{subfigure}\hfill
    \begin{subfigure}[b]{0.25\textwidth}
    \centering
    \includegraphics[trim={0, 0, 0, 0}, clip, width=\textwidth]{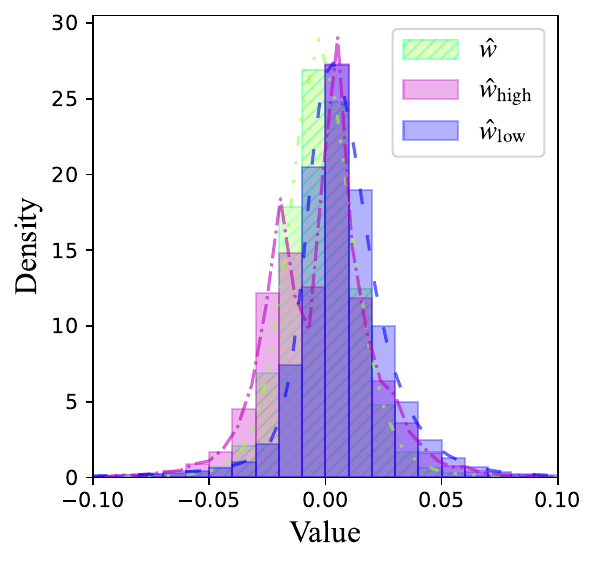}
    \caption{INT(8\textbar2)\label{fig:kde_int8_2}}
    \end{subfigure}\hfill
    \caption{Distribution of $\hat{\bm{w}}$, $\hat{\bm{w}}_\mathrm{high}$ and $\hat{\bm{w}}_\mathrm{low}$. \label{fig:kde_comprasion}}
\end{figure*}
\begin{figure*}[ht]
    \begin{subfigure}[b]{0.25\textwidth}
    \centering
    \includegraphics[trim={0, 0, 0, 0}, clip, width=\textwidth]{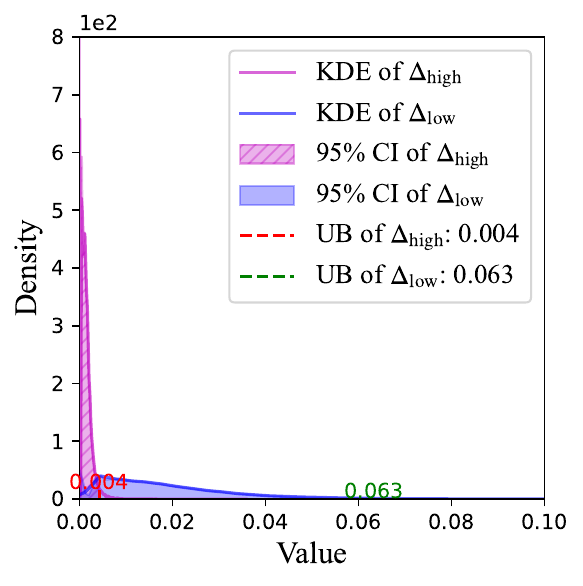}
    \caption{INT(8\textbar5)\label{fig:delta_kde_int8_5}}
    \end{subfigure}\hfill
    \begin{subfigure}[b]{0.25\textwidth}
    \centering
    \includegraphics[trim={0, 0, 0, 0}, clip, width=\textwidth]{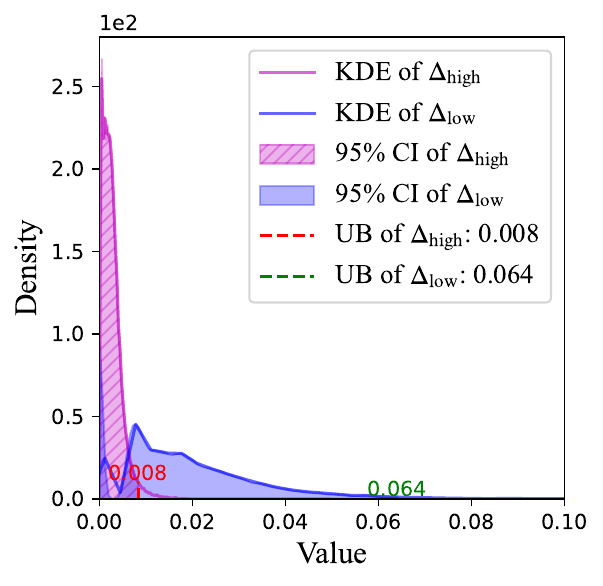}
    \caption{INT(8\textbar4)\label{fig:delta_kde_int8_4}}
    \end{subfigure}\hfill
    \begin{subfigure}[b]{0.25\textwidth}
    \centering
    \includegraphics[trim={0, 0, 0, 0}, clip, width=\textwidth]{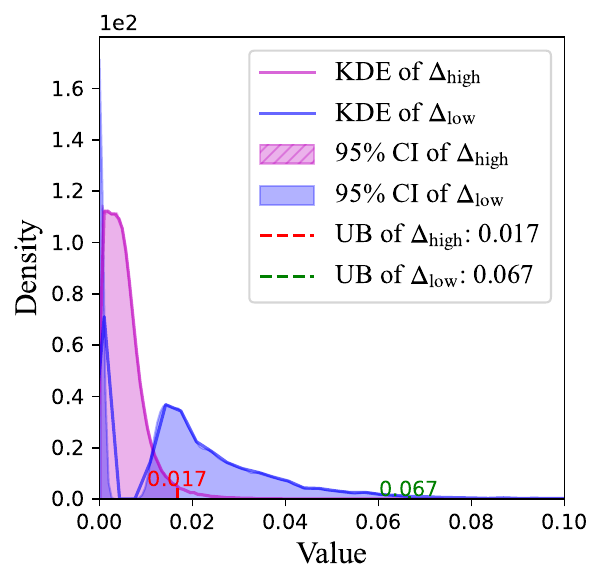}
    \caption{INT(8\textbar3)\label{fig:delta_kde_int8_3}}
    \end{subfigure}\hfill
    \begin{subfigure}[b]{0.25\textwidth}
    \centering
    \includegraphics[trim={0, 0, 0, 0}, clip, width=\textwidth]{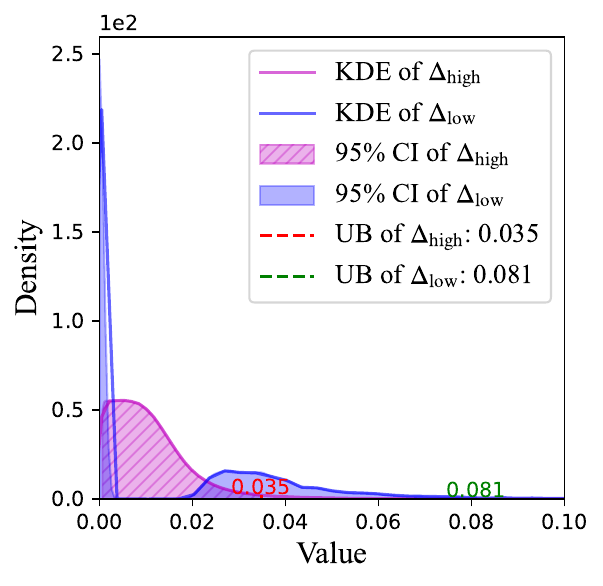}
    \caption{INT(8\textbar2)\label{fig:delta_kde_int8_2}}
    \end{subfigure}\hfill
    \caption{Kernel Density Estimation (KDE) and Upper Bound (UB) of 95\% Confidence Intervals (CIs) of $\Delta_\mathrm{high}$ and $\Delta_\mathrm{low}$. \label{fig:delta_kde_comprasion}}
\end{figure*}

\subsection{NestQuant: Theoretical Foundation}\label{ref:info_anaysis} 
The question is, does \emph{One} data type quantized model hold the potential for diverse models?
In other words, can partial bit weights containing available distributional information be extracted from original quantized weights?

\subsubsection{Integer Weight Decomposition}\label{ref:IWD} 
For the exploration of this, we propose the integer weight decomposition, as shown in Fig.~\ref{fig:iwd}. 
We first specifically use the $\bm{w}_\mathrm{high}$, $\bm{w}_\mathrm{low}$ to denote the higher $h$-bit and lower $l$-bit weights of the $n$-bit weight $\bm{w}_\mathrm{int}$ in quantized model, respectively.
The relation between the higher and lower bits is $n\!=\!h\!+\!l$.
The relation between the higher-bit and lower-bit weight can be defined in the left BitShift:
{
\begin{equation}
\begin{aligned}
\bm{w}_\mathrm{int}&=\mathrm{LeftShift}\left(\bm{w}_\mathrm{high}, l\right) + \bm{w}_\mathrm{low}\\
&=\bm{w}_\mathrm{high} \cdot 2^l + \bm{w}_\mathrm{low}\\
\end{aligned},
\label{eq:decomposition}
\end{equation}
}%
\noindent where the $\mathrm{LeftShift}\left(\bm{x}, l\right)$ is left bitshifting $l$ bits for all numerical values in integer tensors $\bm{x}$. 
So the $n$ bits integer weight $\bm{w}_\mathrm{int}$ can be decomposed into two portions of weight by bit-wise splitting, which is referred to as \emph{higher-bit weight} $\bm{w}_\mathrm{high}$ and \emph{lower-bit weight} $\bm{w}_\mathrm{low}$.
It can also be described in the right BitShift:
{
\begin{equation}
\begin{aligned}
\bm{w}_\mathrm{high}\!&=\mathrm{RightShift}\left(\bm{w}_\mathrm{int}, l\right)\\
&\approx \mathrm{Clip}\left(\left\lfloor \frac{\bm{w}_\mathrm{int}}{2^l} \right\rceil, min_\mathrm{high}, max_\mathrm{high}\right)\ ,\\
\end{aligned}
\label{eq:rightshift}
\end{equation}
}%
\noindent where the $\mathrm{RightShift}\left(\bm{x}, l\right)$ is also right bitshifting $l$ bits for all numerical values in integer tensors $\bm{x}$, and $[min_\mathrm{high}, max_\mathrm{high}]$ is $[-2^{h-1}, 2^{h-1}-1]$.
The transformation from $\bm{w}_\mathrm{int}$ to $\bm{w}_\mathrm{high}$ is actually equivalent to a right bitshifting of $\bm{w}_\mathrm{int}$, and the right bitshifting result $\bm{w}_\mathrm{high}$ can also be approximated as another quantization for $\bm{w}_\mathrm{int}$ with scale $2^l$.
As a result, we can suppose that higher-bit weights $\bm{w}_\mathrm{high}$ implicitly contain some of the available distributional information for the model.
That is, the $\bm{w}_\mathrm{high}$ is also a partial performance contribution of $\bm{w}_\mathrm{int}$.
Then the quantization perturbation between $\bm{w}_\mathrm{int}$ and $\bm{w}_\mathrm{high}$ can be described as:
{
\begin{equation}
\begin{aligned}
\delta \bm{w}_\mathrm{int}\!&=\frac{\bm{w}_\mathrm{int}}{2^l} - \bm{w}_\mathrm{high} = \frac{\bm{w}_\mathrm{low}}{2^l}\\
\end{aligned}.
\label{eq:delta_weight_int}
\end{equation}
}%
\noindent This corresponds to the fact that we have a known quantization perturbation $\delta \bm{w}_\mathrm{int}\!=\!\bm{w}_\mathrm{low}/{2^l}$, and if we {store this known perturbation}, we can switch between two types of different bitwidth and performance models.

The rounding of the $\lfloor \bm{w}_\mathrm{int}/{2^l} \rceil$ is an essential part of the PTQ rounding optimization. 
The forms of Equation~\ref{eq:rightshift} and~\ref{eq:delta_weight_int} are similar to the Hessian-based optimization in Section~\ref{ref:quant_optim_prereq}. 
So, we can approximately optimize the $\bm{w}_\mathrm{high}$ with the Hessian-based algorithm as shown in Equation~\ref{eq:hessian_int} because of the $\delta \bm{w}_\mathrm{int}$ is also a partial contribution of the weights and subject to the Hessian optimization constraints.
{
\begin{equation}
\begin{aligned}
\mathop{\mathrm{argmin}}\limits_{\delta \bm{w}_\mathrm{int}^{(\ell)}}\mathrm{\bf{\mathbb{E}}}\left[ \delta \bm{w}_\mathrm{int}^{(\ell)^\mathrm{T}} \mathbf{H}^{(\bm{w}^{(\ell)}_\mathrm{int})} \delta \bm{w}^{(\ell)}_\mathrm{int} \right]
\end{aligned}.
\label{eq:hessian_int}
\end{equation}
}

Assuming the quantization scale $s$ of $\bm{w}_\mathrm{int}$ is invariant, the dequantization of the higher-bit weight $\bm{w}_\mathrm{int}$ is:
{
\begin{equation}
\begin{aligned}
\hat{\bm{w}}_\mathrm{high}\!=s \cdot 2^l\cdot \bm{w}_\mathrm{high}=s_\mathrm{high}\cdot \bm{w}_\mathrm{high}\\
\end{aligned},
\label{eq:delta_weight_high}
\end{equation}
}%
\noindent where the $\bm{w}_\mathrm{int}$ and $\bm{w}_\mathrm{high}$ share a same scale $s$, for $\bm{w}_\mathrm{high}$ the scale is inflated to $s_\mathrm{high} = s \cdot {2^l}$.

\subsubsection{Similarity Analysis of Decomposed Weights}
To quantitatively analyze the relationships among the decomposed weights, we take the ResNet-18 as an example in the following analysis.
We use the INT($n$\textbar$h$) to represent the INT$n$ weight $\bm{w}_\mathrm{int}$ nesting the INT$h$ weight $\bm{w}_\mathrm{high}$.

We sequentially extract the weights of each layer of the ResNet-18 model (including integer $\bm{w}_\mathrm{int}$, $\bm{w}_\mathrm{high}$, and $\bm{w}_\mathrm{low}$ and dequantized $\hat{\bm{w}}\!=\!s\cdot\bm{w}_\mathrm{int}$, $\hat{\bm{w}}_\mathrm{high}\!=\!s\cdot2^l\cdot\bm{w}_\mathrm{high}$, and $\hat{\bm{w}}_\mathrm{low}\!=\!s\cdot\bm{w}_\mathrm{low}$), flatten and concatenate them into different 1-D vectors. 
The size of these 1-D vectors is all $11,157,504$. 
As a result, we assume that these large vectors can reveal certain statistical properties inherent to the model.
We considered the following metrics from three distinct perspectives:

\vspace{3pt}
\noindent\textbf{Hypothesis Validation.}
We statistically compare distributions using the Wilcoxon Rank-Sum Test in $(\hat{\bm{w}}, \hat{\bm{w}}_\mathrm{high})$ and $(\hat{\bm{w}}, \hat{\bm{w}}_\mathrm{low})$.
If the P-value $p \geq 0.05$, there is no significant difference between the two distributions, whereas the $p < 0.05$, represents the converse result.
As shown in Table~\ref{tab:comparison_wilcox}, the $p$ between $(\hat{\bm{w}}, \hat{\bm{w}}_\mathrm{high})$ is increase from INT(8\textbar2) to INT(8\textbar5).
Specifically, the INT(8\textbar5) and INT(8\textbar4) cases have $p$ of 0.82 and 0.46, respectively, which are all significantly larger than 0.05, representing the similarity of the higher-bit and full-bit weights.
For the $p$ of $(\hat{\bm{w}}, \hat{\bm{w}}_\mathrm{low})$, the $p$ are 0 in all cases.
As shown in Fig.~\ref{fig:kde_comprasion}, the visualizations of weight distributions can also support the results of hypothesis testing.
\begin{table}[ht]
\centering
\caption{Wilcoxon Rank-Sum Test in Nesting ResNet-18.\label{tab:comparison_wilcox}}
\resizebox{.45\textwidth}{!}{%
\begin{tabular}{@{\hspace{2mm}}ccccc@{\hspace{2mm}}}
\toprule
\multirow{2}*{\makecell[l]{Weights Pair}}&
\multicolumn{4}{c}{\makecell[c]{P-value}}
\\ \cmidrule{2-5}
&INT(8\textbar5)&INT(8\textbar4)&INT(8\textbar3)&INT(8\textbar2)\\
\midrule
$(\hat{\bm{w}}, \hat{\bm{w}}_\mathrm{high})$& 0.82 & 0.46 & 0.06 & 0\\ 
$(\hat{\bm{w}}, \hat{\bm{w}}_\mathrm{low})$& 0 & 0 &0 & 0 \\
\bottomrule
\end{tabular}%
}
\end{table}

\noindent\textbf{Confidence Interval Analysis.}
We compute the Kernel Density Estimation (KDE) and 95\% Confidence Intervals (CIs) of the absolute distance $\Delta_\mathrm{high}\!=\!|\hat{\bm{w}}-\hat{\bm{w}}_\mathrm{high}|$ and $\Delta_\mathrm{low}\!=\!|\hat{\bm{w}}-\hat{\bm{w}}_\mathrm{low}|$.
As shown in Fig.~\ref{fig:delta_kde_comprasion}, the lower bounds of $\Delta_\mathrm{high}$ and $\Delta_\mathrm{low}$ in 95\% CIs are almost 0.
We can also observe that the UBs of the 95\% CIs of $\Delta_\mathrm{high}$ decrease from INT(8\textbar2) with 0.035 to INT(8\textbar5) with 0.004, while the UBs of $\Delta_\mathrm{low}$ show no significant change.
The results indicate that the value distances of dequantized higher-bit and full-bit weights approach 0.
\begin{table}[ht]
\centering
\caption{Correlation Test in Nesting ResNet-18.\label{tab:comparison_corr}}
\resizebox{.48\textwidth}{!}{%
\begin{tabular}{@{\hspace{2mm}}lccccc@{\hspace{2mm}}}
\toprule
\multirow{2}*{\makecell[l]{Metric}}&
\multirow{2}*{\makecell[l]{Weights Pair}}&
\multicolumn{4}{c}{\makecell[c]{Correlation}}
\\ \cmidrule{3-6}
&&INT(8\textbar5)&INT(8\textbar4)&INT(8\textbar3)&INT(8\textbar2)\\
\midrule
\multirow{4}*{\makecell[c]{Pearson}}&$(\bm{w}_\mathrm{int}, \bm{w}_\mathrm{high})$& 0.997 & 0.990 & 0.960 & 0.860\\ 
&$(\bm{w}_\mathrm{int}, \bm{w}_\mathrm{low})$& 0.006 & 0.008 &0.009 & -0.015 \\ \cmidrule{2-6}
&$(\hat{\bm{w}}, \hat{\bm{w}}_\mathrm{high})$& 0.995 & 0.980 & 0.925 & 0.774\\ 
&$(\hat{\bm{w}}, \hat{\bm{w}}_\mathrm{low})$& 0.012 & 0.013 & 0.014 & -0.015 \\
\midrule
\multirow{4}*{\makecell[c]{Spearman}}&$(\bm{w}_\mathrm{int}, \bm{w}_\mathrm{high})$& 0.997 & 0.988 & 0.955 & 0.852\\ 
&$(\bm{w}_\mathrm{int}, \bm{w}_\mathrm{low})$& 0.002 & 0.002 & 3.3$e^{-4}$ & -0.042 \\ \cmidrule{2-6}
&$(\hat{\bm{w}}, \hat{\bm{w}}_\mathrm{high})$& 0.992 & 0.972 & 0.905 & 0.761\\ 
&$(\hat{\bm{w}}, \hat{\bm{w}}_\mathrm{low})$& -9.1$e^{-5}$ & 6.7$e^{-4}$ &0.002 & -0.009 \\
\midrule
\multirow{4}*{\makecell[c]{Kendall}}&$(\bm{w}_\mathrm{int}, \bm{w}_\mathrm{high})$& 0.968 & 0.931 & 0.857 & 0.722\\ 
&$(\bm{w}_\mathrm{int}, \bm{w}_\mathrm{low})$& 0.017 & 0.039 &0.082 & 0.143 \\ \cmidrule{2-6}
&$(\hat{\bm{w}}, \hat{\bm{w}}_\mathrm{high})$& 0.929 & 0.861 & 0.738 & 0.566\\ 
&$(\hat{\bm{w}}, \hat{\bm{w}}_\mathrm{low})$& 3.3$e^{-4}$ & 0.003 &0.023 & 0.111 \\
\bottomrule
\end{tabular}%
}
\end{table}

\noindent\textbf{Correlation Analysis.}
The correlations measure the relationship between two variables.
We test the Pearson (linear relationships), Spearman (monotonic relationships), and Kendall (ordinal associations) correlations between the $\bm{w}_\mathrm{int}$, $\bm{w}_\mathrm{high}$, and $\bm{w}_\mathrm{low}$, and dequantized $\hat{\bm{w}}$, $\hat{\bm{w}}_\mathrm{high}$, and $\hat{\bm{w}}_\mathrm{low}$.
As shown in Table~\ref{tab:comparison_corr}, the correlations of $(\bm{w}_\mathrm{int}, \bm{w}_\mathrm{high})$ and $(\hat{\bm{w}}, \hat{\bm{w}}_\mathrm{high})$ are almost larger than 0.9 for Pearson and Spearman, and larger than 0.6 for Kendall.
Thus, the results of higher-bit and full-bit weights correlations suggest that the variables are highly related, either linearly or in terms of rank orders.
The correlations of $(\bm{w}_\mathrm{int}, \bm{w}_\mathrm{low})$ and $(\hat{\bm{w}}, \hat{\bm{w}}_\mathrm{low})$ are all tended to 0.

The results from the three different perspectives all demonstrate the similarity between the $\bm{w}_\mathrm{high}$ and $\bm{w}_\mathrm{int}$, while $\bm{w}_\mathrm{low}$ and $\bm{w}_\mathrm{int}$ are nearly uncorrelated.
Therefore, we can reasonably assume that the $\bm{w}_\mathrm{high}$  retains some of the available distributional information of $\bm{w}_\mathrm{int}$ and can be further leveraged.

\subsubsection{Nesting Mechanism of Decomposed Weights}\label{ref:DW_nest_mech} 
After solving the optimal $\bm{w}_\mathrm{high}$ by Hessian-based adaptive rounding, the lower-bit weight $\bm{w}_\mathrm{low}$ can be calculated by:
{
\begin{equation}
\begin{aligned}
\bm{w}_\mathrm{low}\!=\mathrm{Clip}\left(\bm{w}_\mathrm{int} - \bm{w}_\mathrm{high} \cdot 2^l, max_\mathrm{low}, min_\mathrm{low}\right)\\
\end{aligned},
\label{eq:cal_weight_low}
\end{equation}
}%
\noindent where $[max_\mathrm{low}, min_\mathrm{low}]$ is $[-2^{l-1}, 2^{l-1}-1]$.

In addition to $\bm{w}_\mathrm{high}$ being available as independent weight parameters, the $\bm{w}_\mathrm{high}$ and $\bm{w}_\mathrm{low}$ can be composed to upgrade to the complete bitwidth weight parameters $\bm{w}_\mathrm{int}$.

From $\bm{w}_\mathrm{int}$ to $\bm{w}_\mathrm{high}$ only requires page-out the $\bm{w}_\mathrm{low}$; 
From $\bm{w}_\mathrm{high}$ to $\bm{w}_\mathrm{int}$ only requires page-in the $\bm{w}_\mathrm{low}$ from storage and recomposing, i.e., $\bm{w}_\mathrm{high}$ left bitshifting $l$-bits and then adding with the lower-bit weight $\bm{w}_\mathrm{low}$ in Equation~\ref{eq:decomposition}.
This mechanism is like \emph{nesting} the $\bm{w}_\mathrm{high}$ into $\bm{w}_\mathrm{int}$, we referred it to as \emph{Integer-Nesting Quantization}, i.e., \emph{NestQuant}.

\subsection{NestQuant: Procedures and Switching Instances} \label{ref:nestquant_framework}

\begin {figure}[t]
\centering
\includegraphics[width=0.95\linewidth]{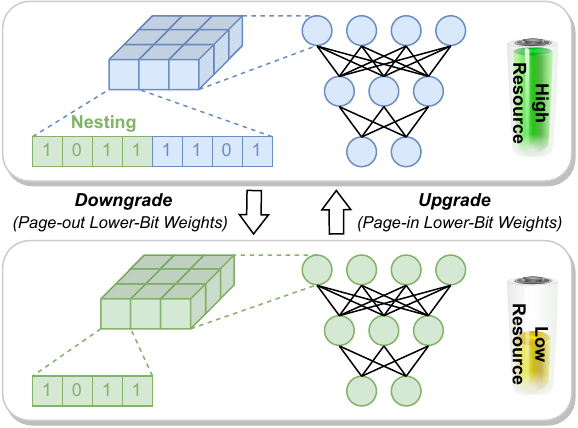}
\caption {Overview of the proposed NestQuant.} 
\label{fig:overview}
\end{figure}

The overview of our proposed NestQuant is shown in Fig.~\ref{fig:overview}.
The quantized INT$n$ model, referred to as \emph{full-bit model}, nests a quantized INT$h$ model, which is referred to as \emph{part-bit model}. 
The NestQuant can achieve on-device switching between two quantized models by simply storing the $\bm{w}_\mathrm{high}$ and $\bm{w}_\mathrm{low}$ for resource adaptation.

The full-bit model can be launched in adequate resource conditions on devices. 
When resources are insufficient, the full-bit model can downgrade to the part-bit model by paging out the $\bm{w}_\mathrm{low}$ to adapt to less resource requirements.
The part-bit model can page in the $\bm{w}_\mathrm{low}$ from storage to upgrade to the high-performing full-bit model when resources become available. 

We designate the rounding optimization of NestQuant based on the data-free Hessian-based PTQ adaptive rounding algorithm, SQuant\cite{guo2022squant}.
The layer-wise procedures of the proposed NestQuant are as shown in Algorithm~\ref{alg:nestquant_procs}:
\begin{algorithm}
\caption{NestQuant Procedures}
\label{alg:nestquant_procs}
\KwIn{A floating-point DNN model ${\bf{\mathcal{M}}}_{fp}$}
\KwOut{An INT($n$\textbar$h$) model ${\bf{\mathcal{M}}}_{\mathrm{int}(n|h)}$}

\For{weights $\bm{w}^{(\ell)}$ of layer $1,2, \ldots, \ell$ in ${\bf{\mathcal{M}}}_{fp}$}{
    \addtocounter{algocf}{-1}
    \textcolor{gray}{/* Step \ding{172}: INT$n$ Hessian-based quantization for $\bm{w}_\mathrm{int}^{(\ell)}$ */}
    \addtocounter{algocf}{1} 

    \hspace*{5pt} Compute scale $s^{(\ell)}$ by linear quantization\; 

    \hspace*{5pt} Compute floating-point weights $\frac{\bm{w}^{(\ell)}}{s^{(\ell)}}$\;

    \hspace*{5pt} Optimize integer data type $\bm{w}_\mathrm{int}^{(\ell)}$ by adaptive \hspace*{5pt} rounding in $\left\lfloor\frac{\bm{w}^{(\ell)}}{s^{(\ell)}}\right\rceil$ and clipping\;

    \textcolor{gray}{/* Step \ding{173}: INT$h$ Hessian-based quantization for $\bm{w}_\mathrm{high}^{(\ell)}$ */} \Comment{Section~\ref{ref:IWD} \& \ref{ref:DW_nest_mech}}

    \hspace*{5pt} {Select effective nested bits $h$;} \Comment{Section~\ref{ref:sweetspot}}
    
    \hspace*{5pt} {Compute lower bits $l = n - h$}\;
    
    \hspace*{5pt} Store nesting scale $s_\mathrm{high}^{(\ell)} = s^{(\ell)} \cdot 2^l$\;
    
    \hspace*{5pt} Compute floating-point nesting weights $\frac{\bm{w}_\mathrm{int}^{(\ell)}}{2^l}$\;

    \hspace*{5pt} {Optimize integer data type $\bm{w}_\mathrm{high}^{(\ell)}$ by adaptive \hspace*{5pt} rounding in $\left\lfloor \frac{\bm{w}_\mathrm{int}^{(\ell)}}{2^l} \right\rceil$ and clipping}\;

    \hspace*{5pt} Compute $\bm{w}_\mathrm{low}^{(\ell)}$ by $\bm{w}_\mathrm{int}^{(\ell)} - \bm{w}_\mathrm{high}^{(\ell)} \cdot 2^l$\;

    \hspace*{5pt} {Extra 1-bit range of clipping for $\bm{w}_\mathrm{low}^{(\ell)}$ for \hspace*{5pt} compensation}; \Comment{Section~\ref{ref:compen}}

    \textcolor{gray}{/* Step \ding{174}: Pack and store $h$-bit and ($l$+1)-bit weights */}
    
    \hspace*{5pt} $\bm{w}^{(\ell)}\longleftarrow \{\bm{w}_\mathrm{high}^{(\ell)}, \bm{w}_\mathrm{low}^{(\ell)}\}$\
}
\textbf{Return} ${\bf{\mathcal{M}}}_{\mathrm{int}(n|h)}$
\end{algorithm}

For the each layer $\ell$ of FP32 model ${\bf{\mathcal{M}}}_{fp}$, we make a layer-wise weights nesting for $\bm{w}^{(\ell)}$:

\noindent (i) The first step is processing a Hessian-based quantization for $\bm{w}_\mathrm{int}^{(\ell)}$, which includes computing the $s^{(\ell)}$, ${\bm{w}^{(\ell)}}/{s^{(\ell)}}$ and deriving the $\bm{w}_\mathrm{int}^{(\ell)}$ from the adaptive rounding with $\left\lfloor{\bm{w}^{(\ell)}}/{s^{(\ell)}}\right\rceil$.
Then, selecting the effective nested bits $h$ by the pattern of critical nested combination.

\noindent (ii) The second step is decomposing the $\bm{w}_\mathrm{int}^{(\ell)}$ from step-(i) and has a secondary Hessian-based quantization for $\bm{w}_\mathrm{high}^{(\ell)}$. That is, calculating the lower bits $l = n - h$ and nesting scale $s_\mathrm{high}^{(\ell)}$ to determine the nesting bitshifting, and computing the ${\bm{w}_\mathrm{int}^{(\ell)}}/{2^l}$ and deriving the $\bm{w}_\mathrm{high}^{(\ell)}$ from the adaptive rounding with $\left\lfloor {\bm{w}_\mathrm{int}^{(\ell)}}/{2^l} \right\rceil$.
After determining the optimal $\bm{w}_\mathrm{high}^{(\ell)}$, the $\bm{w}_\mathrm{low}^{(\ell)}$ can also be computed by $\bm{w}_\mathrm{int}^{(\ell)} - \bm{w}_\mathrm{high}^{(\ell)} \cdot 2^l$ and also with an extra 1-bit range compensation.

\noindent (iii) On the third step, the $h$-bit $\bm{w}_\mathrm{high}^{(\ell)}$ and ($l$+1)-bit $\bm{w}_\mathrm{low}^{(\ell)}$ are registered in the model to replace the original FP32 weight $\bm{w}^{(\ell)}$ in layer $\ell$.

Repeating the above layer-wise procedures for all layers in ${\bf{\mathcal{M}}}_{fp}$, we can obtain the NestQuant model ${\bf{\mathcal{M}}}_{\mathrm{int}(n|h)}$, which can switch between full-bit and part-bit models.

\subsubsection{Part-Bit Model: Effective Nested Combinations} \label{ref:sweetspot} 

\begin {figure}[t]
\centering
\includegraphics[width=0.85\linewidth]{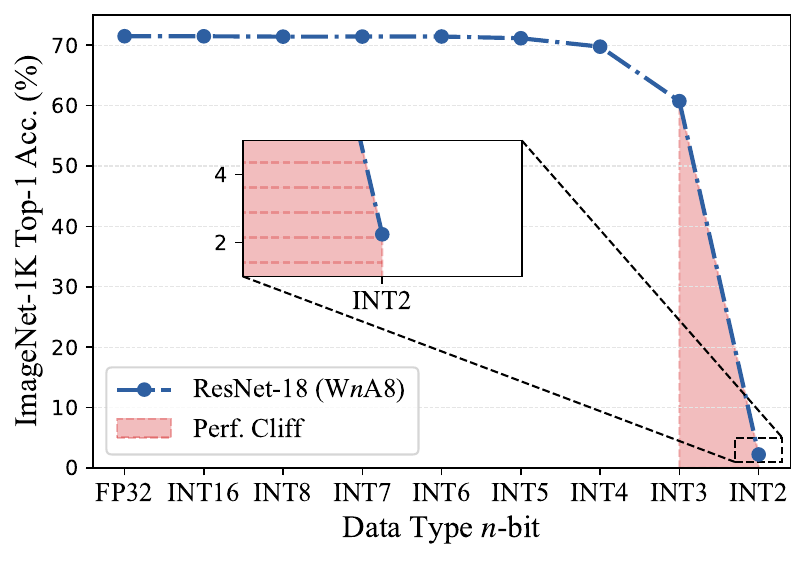}
\caption {Performance cliff of the PTQ model.} 
\label{fig:perfcliff}
\end{figure}

As shown in Fig.~\ref{fig:perfcliff}, the performance of the model with common bits weight, e.g., INT8 quantized model, is similar to the FP32 model performance, but there is an extreme performance drop in INT2.
In this paper, we refer to the large performance decrease in lower bits as \emph{Performance Cliff} in quantization.

For the part-bit model, it is important to determine the feasible combinations of bits $n$ and nested bits $h$, which are referred to as \emph{Effective Nested Combinations}.
We also name the effective nested combination before the performance cliff as the \emph{Critical Nested Combination}.
8-bit and 6-bit weights can nest lower bitwidth types, while models below 4-bit are too ill-performing to nest due to the performance cliff.
Most 2-bit and 1-bit models are barely usable in PTQ, these data types would not be considered for nesting.
As a result, we consider the nesting combination experiments of INT(8\textbar$h$), $h_{\in \{3,4,...,7\}}$ and INT(6\textbar$h$), $h_{\in \{3,4,5\}}$.
\begin{table}[ht]
\centering
\caption{INT8 Nesting Test in ResNet-18.\label{tab:integer_nest}}
\resizebox{.48\textwidth}{!}{%
\begin{tabular}{@{\hspace{2mm}}llcccc@{\hspace{2mm}}}
\toprule
\multirow{3}*{\makecell[l]{Method}}
&\multirow{3}*{\makecell[l]{A-bit}}
&\multirow{3}*{\makecell[c]{W-bit}}
&\multicolumn{3}{c}{ImageNet-1K Top-1 Acc. (\%)}\\ \cmidrule{4-6}
&&& \multirow{2}*{\makecell[c]{Part-Bit}} 
& \multirow{2}*{\makecell[c]{Full-Bit\\(w/o compen.)}}
& \multirow{2}*{\makecell[c]{Full-Bit}} 
\\
\\ \midrule
- & FP32 & FP32 & - & - & 71.5 \\ 
- & INT8 & INT8 & - & - & 71.4 \\ \midrule
BitShift & INT8 & INT(8\textbar4) & 0.1 & 0.1 & 71.4 \\ 
RTN & INT8 & INT(8\textbar4) & 14.5 & 69.7 & 71.4 \\  \midrule
\multirow{5}*{\makecell[l]{Adaptive\\Rounding}}&\multirow{5}*{\makecell[l]{INT8}}
& INT(8\textbar3) & 23.8 & 48.4 & 71.4 \\ 
&& INT(8\textbar4) & 64.1 & 68.7 & 71.4 \\ 
&& INT(8\textbar5) & 70.6 & 69.0 & 71.4 \\
&& INT(8\textbar6) & 71.2 & 65.9 & 71.4 \\ 
&& INT(8\textbar7) & 71.4 & 49.9 & 71.4 \\ \bottomrule
\end{tabular}%
}
\end{table}

In the part-bit model, the utilization of adaptive rounding significantly impacts performance.
As shown in Table~\ref{tab:integer_nest}, the INT(8\textbar4) part-bit performances of utilizing BitShift and RTN are so low as to be unusable, whereas adaptive rounding can maintain a good accuracy.

\vspace{3pt}
\noindent\textbf{Model-Dependence.} Unfortunately, \emph{effective nested combinations} (e.g., INT($n$\textbar$h$), where $n\!=\!8, h \in \{4,...,7\}$) and the \emph{critical nested combination} (i.e., INT($n\!=\!8$\textbar$h\!=\!4$)) of NestQuant are model dependent, as shown in the following evaluations of representative DNNs in Section~\ref{ref:4-2-perf_dnn} and Section~\ref{ref:vit-experiment}.
For example, the effective nested combinations of lightweight CNNs require \emph{higher nested bits} (e.g., $h\!=\!5$), standard CNNs require \emph{lower nested bits} (e.g., $h\!=\!4$), and ViTs require \emph{even lower nested bits} (e.g., $h\!=\!3$).
Given the above observation, we hypothesize that the effective nested combinations of a DNN are related to the capacity of the model, i.e., model parameters or size.

\begin {figure}[t]
\centering
\includegraphics[width=0.65\linewidth]{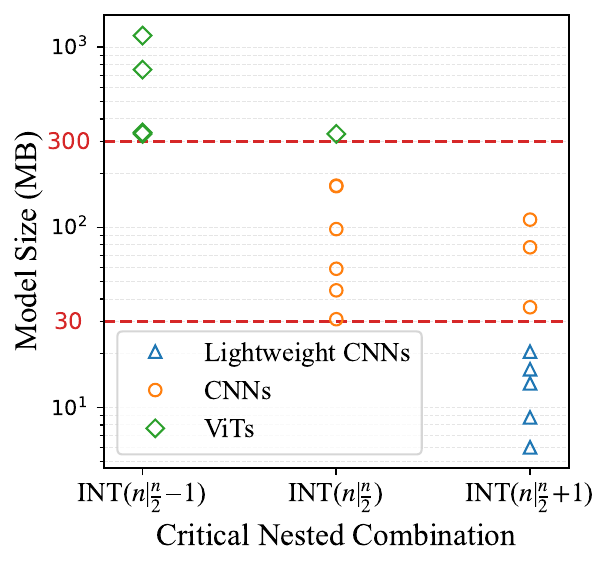}
\caption {Critical nested combination-Model size relationships.} 
\label{fig:robust_nest}
\end{figure}

\vspace{3pt}
\noindent\textbf{Emerging Patterns.} To validate our hypothesis on the potential linkage between model capacity and its effective nesting, we visualize the ``Critical nested combination vs. Model size'' relationship in INT8 nesting, as shown in Fig.~\ref{fig:robust_nest}.
Evidently, two clear cut-offs at model sizes of 30 MB and 300 MB emerged.
Accordingly, given FP32 model size $D_\mathrm{fp32}$, we can identify the following patterns of the critical nested combination:
{
\begin{equation}
\left\{
\begin{aligned}
&\mathrm{INT}\!\left(n|{n}/{2}\!+\!1\right), &&\mathrm{if}\ 0 <D_\mathrm{fp32}<3\!\cdot\!10^1 \mathrm{MB},\\
&\mathrm{INT}\!\left(n|{n}/{2}\right), &&\mathrm{if}\ 3\!\cdot\!10^1\mathrm{MB}\le D_\mathrm{fp32} < 3\!\cdot\!10^2\mathrm{MB},\\ 
&\mathrm{INT}\!\left(n|{n}/{2}\!-\!1\right), &&\mathrm{if}\ D_\mathrm{fp32} \ge 3\!\cdot\!10^2\mathrm{MB}.
\end{aligned}\right.
\label{eq:partbit_pattern}
\end{equation}
}%
\noindent Once the critical nested bit is identified, the effective nested combinations can also be determined.
For simplicity, adopting INT($n$\textbar$\frac{n}{2}\!+\!1$) for CNNs and INT($n$\textbar$\frac{n}{2}$) for ViTs can usually yield the effective  NestQuant model in INT8 nesting.

\subsubsection{Full-Bit Model: Performance Compensation} \label{ref:compen} 
\begin {figure}[t]
\centering
\includegraphics[width=0.9\linewidth]{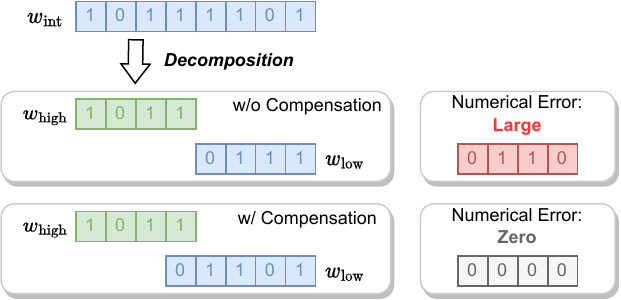}
\caption {
The case of decomposed signed INT data with numerical errors. 
The $\bm{w}_\mathrm{high}$ is $-5$ in BitShift, i.e., 1011 (in 2's complement code).
The $\bm{w}_\mathrm{low}$ is $\mathrm{Clip}(-67 – (-5)\cdot2^4, -2^{4-1}, 2^{4-1}-1) = 7$, i.e., 0111.
The recomposed data $\bm{w}^\mathrm{recomp}_\mathrm{int} = (-5) \cdot 2^4 + 7 = -73$,  i.e., 1011 0111.
The numerical error is $\bm{w}_\mathrm{int} - \bm{w}^\mathrm{recomp}_\mathrm{int} = (-67) – (-73) = 6$.
If we compensate an extra 1-bit for $\bm{w}_\mathrm{low}$ to store the 01101, i.e., $13$, the recomposed results will have no numerical errors.
} 
\label{fig:exta1bit}
\end{figure}

For the full-bit model, it is also important to retain the original performance when recomposing the weights.
However, the decomposed rounding/bitshifting of the $\bm{w}_\mathrm{int}$ may result in a change of $\bm{w}_\mathrm{high}$.
In this case, the range clipping of lower-bit weight within $[-2^{l-1}, 2^{l-1}-1]$ in Equation~\ref{eq:cal_weight_low} will lose information.
These changes will introduce \emph{Numerical Errors}, as shown in Fig.~\ref{fig:exta1bit}.
The bits range of $\bm{w}_\mathrm{low}$ will significantly affect the performance of the recomposed $\bm{w}_\mathrm{int}$ due to numerical errors as depicted in Table~\ref{tab:integer_nest}.
To quantitatively analyze the causes of errors, we calculate the errors generated during the decomposing-recomposing process for all numbers within the range $[-128, 127]$ in BitShift, RTN, Rounding Up and Down (the adaptive rounding is a type of mixed Rounding Up and Down), as shown in Table~\ref{tab:comparison_numerr}.
We can observe that the ranges of errors all lie within $[-2^{l-1}+1, 2^{l-1}]$ for different types of INT($n$\textbar$h$) nesting.
The result of adding the numerical error's range $[-2^{l-1}+1, 2^{l-1}]$ to the clipped $\bm{w}_\mathrm{low}$'s range $[-2^{l-1}, 2^{l-1} - 1]$ is $[-2^l+1, 2^l - 1]$, which is precisely contained within the signed INT($l$+1) range $[-2^l, 2^l - 1]$.
\begin{table}[ht]
\centering
\caption{Nesting Numerical Errors of Signed INT8 Numbers from -128 to 127 (Total 256).\label{tab:comparison_numerr}}
\resizebox{.48\textwidth}{!}{%
\begin{threeparttable}[l]
\begin{tabular}{@{\hspace{2mm}}llccccc@{\hspace{2mm}}}
\toprule
\multirow{2}*{\makecell[l]{Method}}&
\multirow{2}*{\makecell[l]{Metrics of\\Num. Err.}}&
\multicolumn{5}{c}{\makecell[c]{Nesting Numerical Errors}}
\\ \cmidrule{3-7}
&&INT(8\textbar7)
&INT(8\textbar6)
&INT(8\textbar5)
&INT(8\textbar4)
&INT(8\textbar3)\\
\midrule
\multirow{2}*{\makecell[l]{BitShift}}
&\#Non-zero\tnote{(1)}& 128& 128 & 128 & 128 & 128 \\  
&Range& $[0,1]$ & $[0,2]$ & $[0,4]$ & $[0,8]$ & $[0,16]$\\
\midrule
\multirow{2}*{\makecell[l]{RTN}}
&\#Non-zero& 65 & 34 & 20 & 16 & 20\\ 
&Range& $[0,1]$ & $[0,2]$ & $[0,4]$ & $[0,8]$ & $[0,16]$\\
\midrule
\multirow{2}*{\makecell[l]{Rounding\\Up}}
&\#Non-zero& 1 & 65 & 97 & 113 & 121\\ 
&Range& $[0,1]$ & $[-1,2]$ & $[-3,4]$ & $[-7,8]$ & $[-15,16]$\\
\midrule
\multirow{2}*{\makecell[l]{Rounding\\Down}}
&\#Non-zero& 128& 128 & 128 & 128 & 128 \\  
&Range& $[0,1]$ & $[0,2]$ & $[0,4]$ & $[0,8]$ & $[0,16]$\\
\bottomrule
\end{tabular}%

\begin{tablenotes}
\footnotesize
\item[(1)] \emph{The number of non-zero values (\#Non-zero) in numerical errors represents the number of errors generated in recomposing without compensation.} 
\end{tablenotes}
\end{threeparttable}
}
\end{table}

\begin {figure}[t]
\centering
\includegraphics[width=0.85\linewidth]{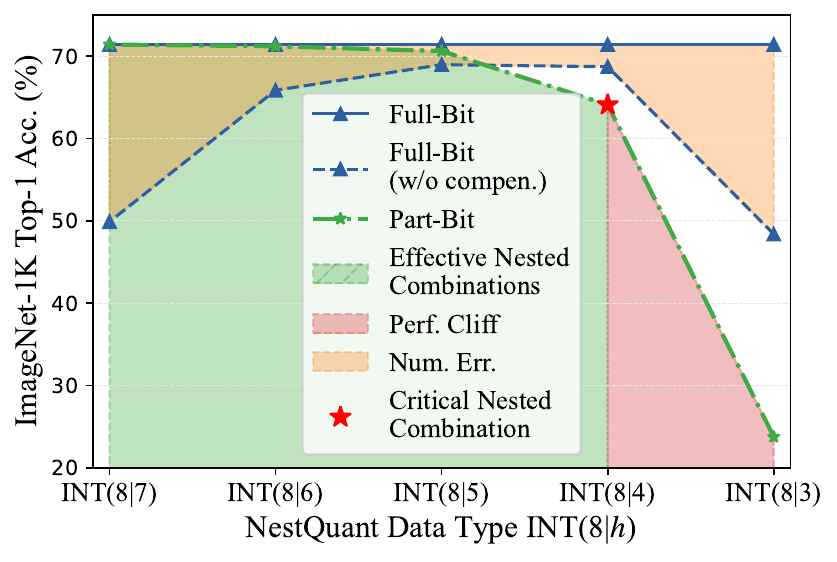}
\caption {NestQuant effective nested combinations, compensation, and critical nested combination.}
\label{fig:sweetspot}
\end{figure}

Therefore, we introduce an extra 1-bit compensation for all $\bm{w}_\mathrm{low}$, which is performed for all lower-bit weights all the time with ($l$+1)-bit range for $\bm{w}_\mathrm{low}$ to expand the range of numerical values.

%

\subsubsection{NestQuant Implementation} \label{ref:nestquant_engine}
\textbf{Model Selection.} Because the part-bit model of critical nested combination has available performance and a sufficiently lower bit compared to the full-bit one, it is more suitable for practical deployment than other effective nested combinations.
For example, for monitoring cameras powered by solar energy/battery providing intelligent detection services, a full-bit INT8 model can be used in busy scenarios to obtain better accuracies, while the part-bit INT4 model in critical nested combination can be used in insignificant scenarios to further save resources.

\vspace{3pt}
\noindent\textbf{Deployment.}
As described in Section~\ref{ref:2-limit}, the representative on-device DL libraries currently do not support arbitrary $k$-bit data types for the implementation of quantized DNN models.
We employ the Packed-Bit Tensors algorithm from works~\cite{petersen2022difflogic, petersen2023distributional} to pack a FP32 tensor into an INT64 tensor in NestQuant. 
The FP32 tensor's data are constrained to the range of signed INT$k$ values $[-2^{k-1}, 2^{k-1} -1]$, where $k_{\in \{3,4,... ,7\}}$. 
The algorithm compresses $64//k$ $k$-bit data elements into a single unsigned INT64 value, where $64//k$ represents the number of packed elements. 
For example, an unsigned INT64 weight can pack twenty-one 3-bit weights, and can also pack twelve 5-bit weights, etc.

\vspace{3pt}
\noindent\textbf{Storing \& Scheduling.}
For the storing and scheduling in deployment, we only need to store the INT$h$ $\bm{w}_\mathrm{high}$ and INT($l$+1) $\bm{w}_\mathrm{low}$ separately in packed-bit weights.
Storing with 1-bit compensation can not utilize the BitShift for switching, it is a compromise for compensating model performance. 
For launching the part-bit model, we can load the $\bm{w}_\mathrm{high}$ in the model file (e.g., \emph{.pth}) by the variable name of $\bm{w}_\mathrm{high}$.
For upgrading the part-bit model to the full-bit model, we can load the $\bm{w}_\mathrm{low}$ by its variable name in the model file and recompose with $\bm{w}_\mathrm{high}$ at runtime for inference.
So the NestQaunt model will maintain two integer weights in the model file in deployment.

The compensation method is relatively straightforward but effective.
This will introduce extra storage space to store the $\bm{w}_\mathrm{low}$, but the cost is acceptable as the evaluated results shown in Section~\ref{ref:modelsize}, which is a trade-off and like extra \emph{One-bit} to swap \emph{One different model}.

\section{Experiments} \label{ref:4-experiment} 

\begin{figure*}[ht]
    \begin{subfigure}[b]{0.33\textwidth}
    \centering
    \includegraphics[trim={0, 0, 0, 0}, clip, width=\textwidth]{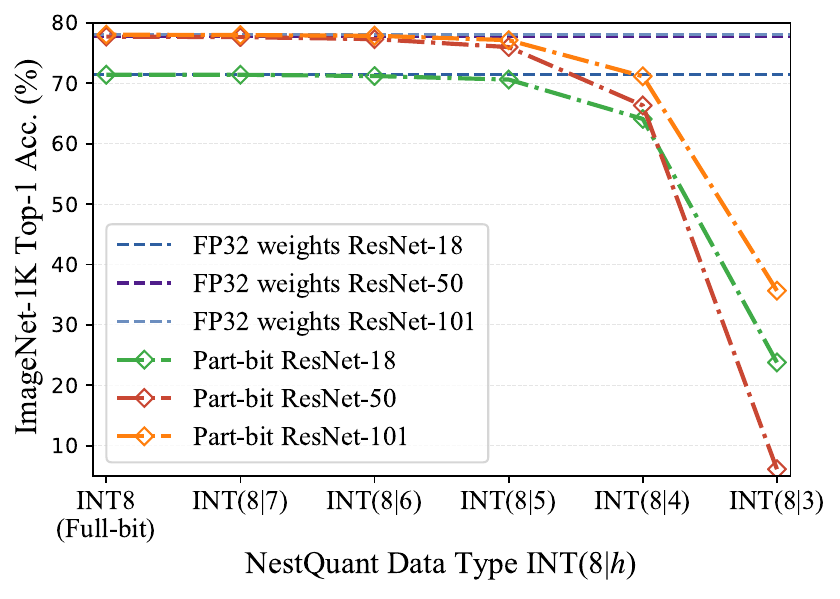}
    \caption{ResNet\label{fig:rs_int8_nesting}}
    \end{subfigure}\hfill
    \begin{subfigure}[b]{0.33\textwidth}
    \centering
    \includegraphics[trim={0, 0, 0, 0}, clip, width=\textwidth]{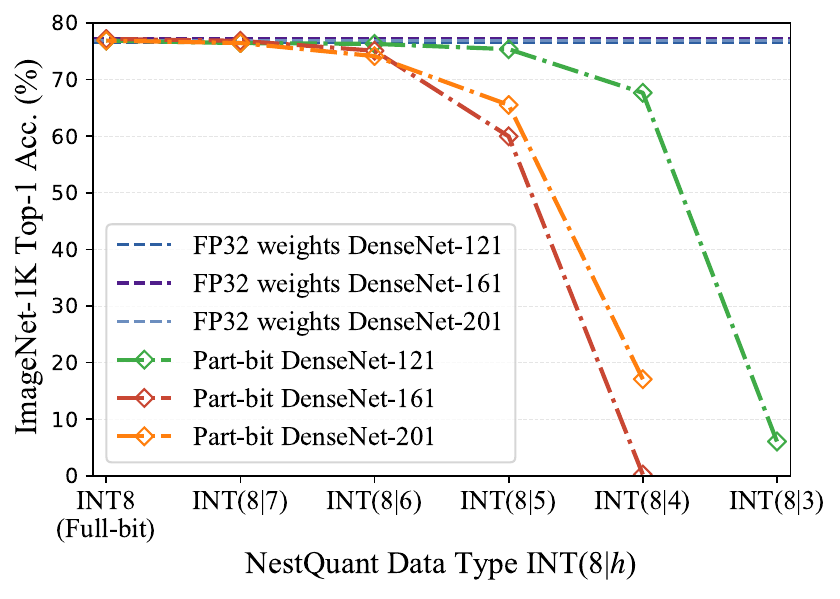}
    \caption{DenseNet\label{fig:ds_int8_nesting}}
    \end{subfigure}\hfill
    \begin{subfigure}[b]{0.33\textwidth}
    \centering
    \includegraphics[trim={0, 0, 0, 0}, clip, width=\textwidth]{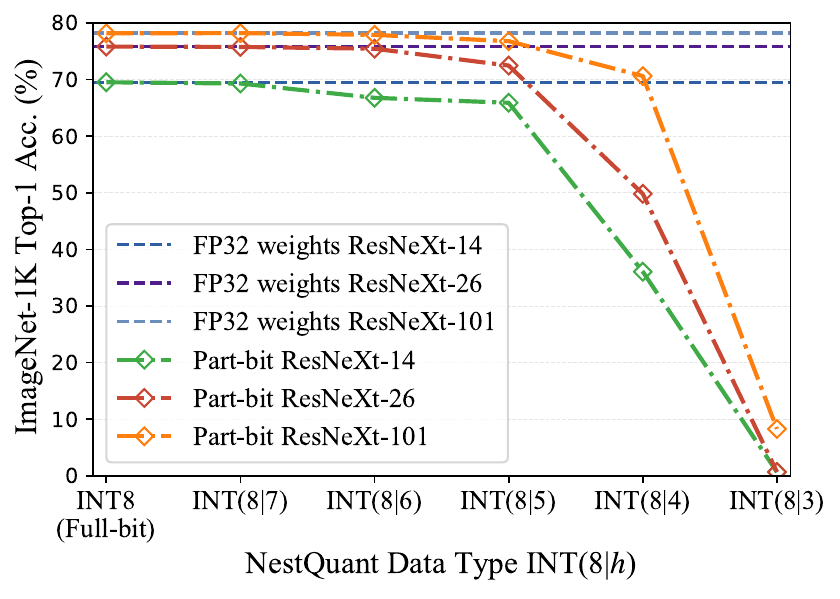}
    \caption{ResNeXt (32x4d)\label{fig:rsx_int8_nesting}}
    \end{subfigure}\hfill
    \caption{Performance of INT8 nesting quantization. \label{fig:int8_nesting}}
\end{figure*}

\begin{figure*}[ht]
    \begin{subfigure}[b]{0.33\textwidth}
    \centering
    \includegraphics[trim={0, 0, 0, 0}, clip, width=\textwidth]{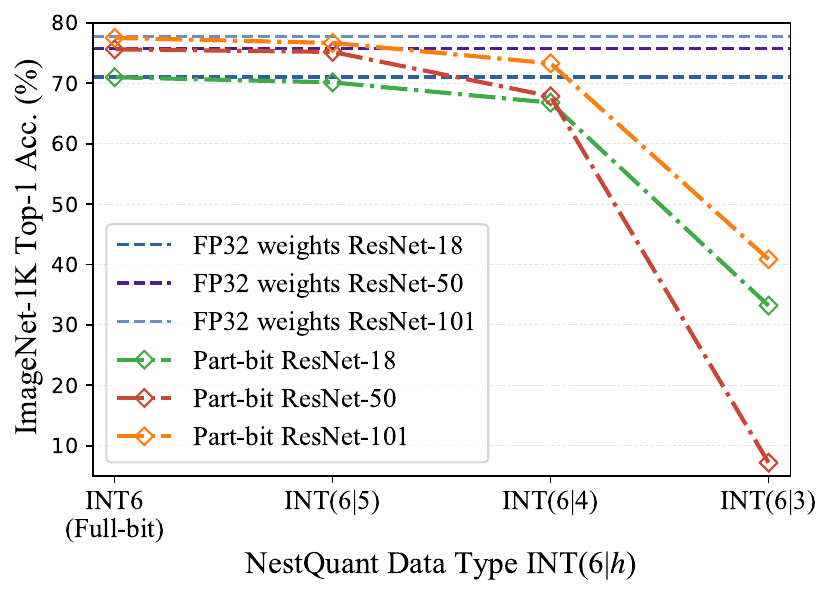}
    \caption{ResNet\label{fig:rs_int6_nesting}}
    \end{subfigure}\hfill
    \begin{subfigure}[b]{0.33\textwidth}
    \centering
    \includegraphics[trim={0, 0, 0, 0}, clip, width=\textwidth]{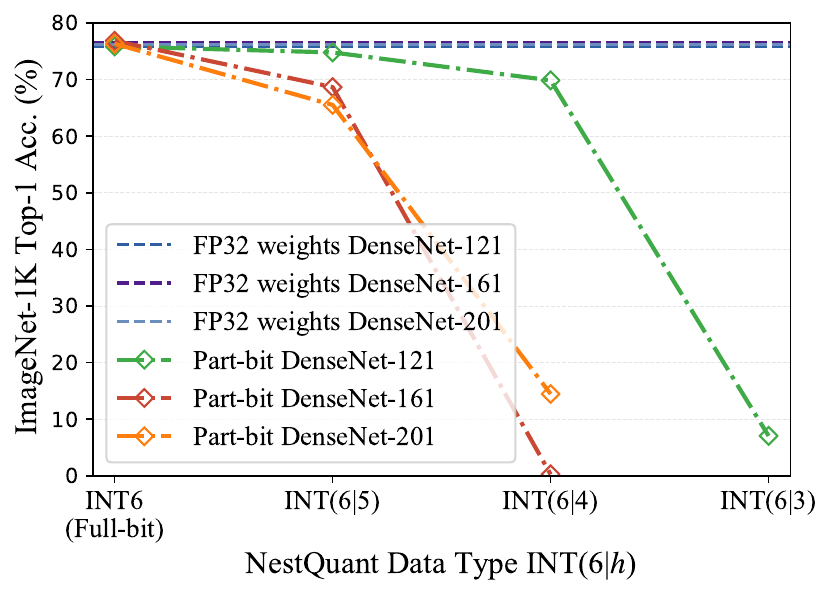}
    \caption{DenseNet\label{fig:ds_int6_nesting}}
    \end{subfigure}\hfill
    \begin{subfigure}[b]{0.33\textwidth}
    \centering
    \includegraphics[trim={0, 0, 0, 0}, clip, width=\textwidth]{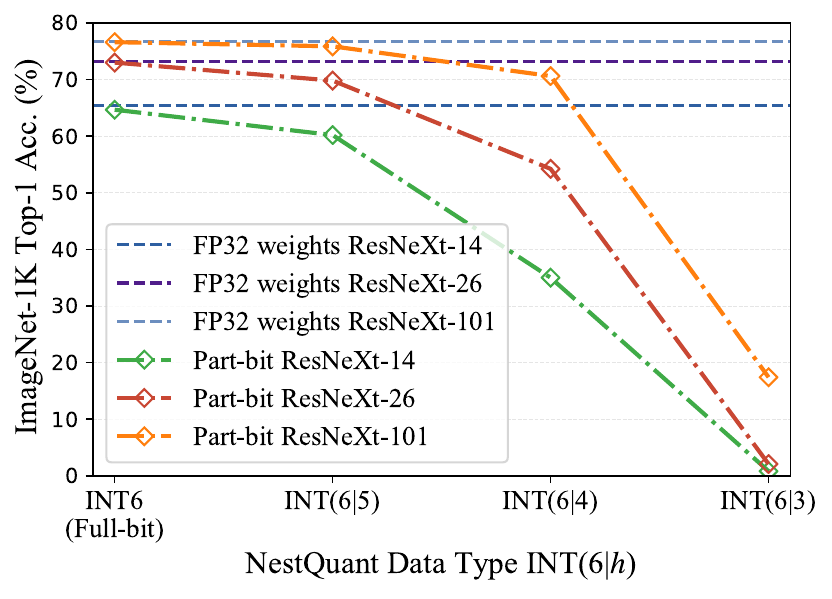}
    \caption{ResNeXt (32x4d)\label{fig:rsx_int6_nesting}}
    \end{subfigure}\hfill
    \caption{Performance of INT6 nesting quantization. \label{fig:int6_nesting}}
\end{figure*}

\begin {figure}[t]
\centering
\includegraphics[width=1.0\linewidth]{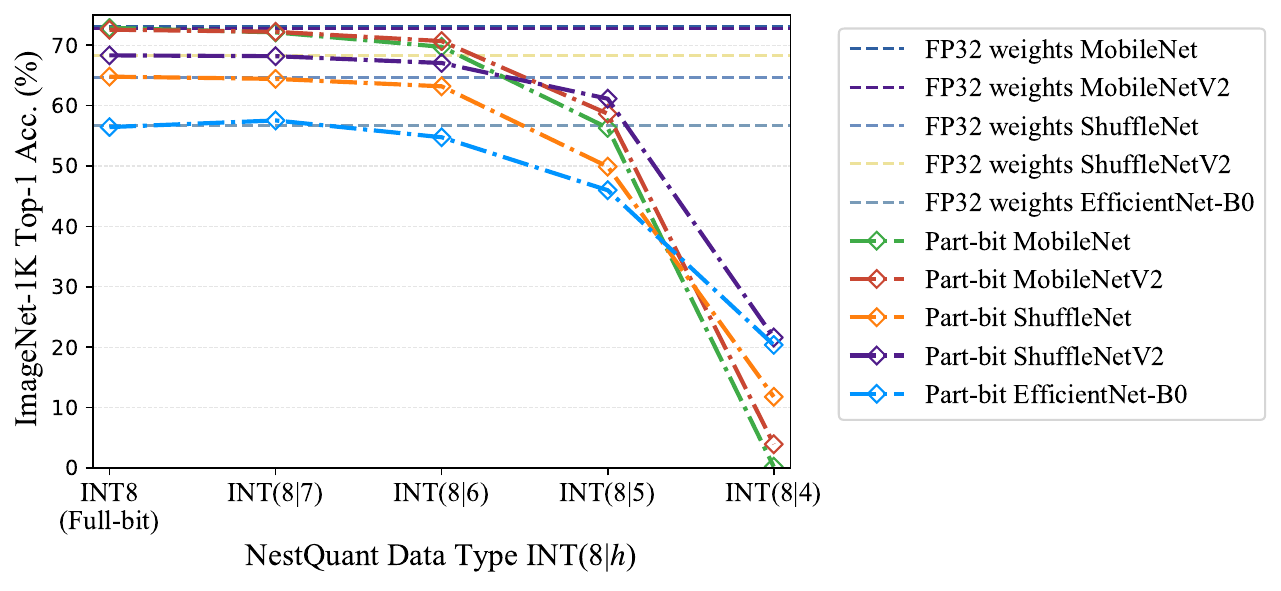}
\caption {Performance of INT8 nesting in lightweight CNNs.} 
\label{fig:lw_int8_nesting}
\end{figure}

\subsection{Experimental Setup}
\noindent\textbf{Datasets.} The widely used computer vision dataset for DL vision services is considered for evaluating the efficacy of our proposed model quantization technology.

{\emph{ImageNet}}~\cite{deng2009imagenet} is a cornerstone dataset in evaluating the DNNs performance of computer vision, and a baseline for pretrained DNN models.
ImageNet-1K is a subset of ILSVRC ImageNet, which contains 1.28 million training images and 50K validation images from 1K different classes.

For post-training compression, related works are usually utilizing the ImageNet pretrained DNNs as the baseline FP32 models, and in terms of PTQ, the PTQ algorithms often use a subset of training images from ImageNet-1K for calibration.
Because the proposed NestQuant is verified with the data-free algorithm SQuant~\cite{guo2022squant}, we only use the ImageNet-1K validation set for the performance evaluation without calibration.

\noindent\textbf{Baseline DNN models.} Since pretrained models are often the basis for fine-tuning other on-device AI services, we validate the effectiveness of the NestQuant by the performance of nesting quantized pretrained models.
We considered the representative and widely-used CNN models~\cite{He@Deep, huang2017densely, xie2017Aggregated, Howard@MobileNets, Sandler@MobileNetV2, Zhang@ShuffleNet, Ma@ShuffleNetV2, tan2019efficientnet} as the baseline, including the ResNet~\cite{He@Deep}, DenseNet~\cite{huang2017densely}, ResNeXt~\cite{xie2017Aggregated}, and lightweight CNN models like MobileNet~\cite{Howard@MobileNets}, MobileNetV2~\cite{Sandler@MobileNetV2}, ShuffleNet~\cite{Zhang@ShuffleNet}, ShuffleNetV2~\cite{Ma@ShuffleNetV2}, and EfficientNet~\cite{tan2019efficientnet}, which are all ImageNet pretrained DNN models in PyTorchCV Library and have a measurable standard baseline.
We also considered the larger ViT models including ImageNet-1K pretrained ViT~\cite{Dosovitskiy@vit}, DeiT~\cite{touvron21deit}, and Swin~\cite{liu2021swin}.

\noindent\textbf{Evaluation Metrics.} We evaluated the performance of the NestQuant model in two aspects: (i) ImageNet-1K Top-1 accuracy for classification problems as the performance, and (ii) on-device resource measurement, contains network traffic in data transmission, storage size and switch cost in memory for IoT devices.

\noindent\textbf{Implementation Details.} We carried out the post-training optimization experiments in PyTorch 1.10 with CUDA 11, performed on NVIDIA
RTX 2080Ti GPUs server.
For the deployment of NestQuant, we used the IoT development board of Raspberry Pi 4B as the IoT device, which CPU is Broadcom BCM2711 with 4GB RAM, and the Wi-Fi link protocol for transmission is IEEE 802.11ac (2.4/5GHz).

\subsection{Performance Evaluation}\label{ref:4-2-perf_dnn}
In this subsection, we will describe the effective nested combinations of NestQuant models.
All models are set with extra 1-bit compensation for lower-bit weights.
With the compensation, all the part-bit models can upgrade to the full-bit ones without any performance degradation.
For a fair comparison, we set up the 8-bit activations in INT8 nesting, and 6-bit activations in INT6 nesting, and the bitwidth of the weights is nested.
\subsubsection{Evaluation in Representative CNN Models}\label{ref:perf_evaluation_cnn}
\noindent\textbf{ResNet.} 
For ResNet series (ResNet-18/-50/-101) with INT(8\textbar$h$), $h_{\in \{3,4,5,6,7\}}$, as shown in Fig.~\ref{fig:rs_int8_nesting}, compared to FP32 weights models with accuracies of 71.5\%, 77.7\% and 78.1\%, and full-bit models with accuracies of 71.4\%, 77.7\% and 78.1\%, the part-bit models of INT(8\textbar$h$), $h_{\in \{5,6,7\}}$ exhibit negligible performance degradation.
The accuracies of INT(8\textbar4) models are 64.1\%, 66.3\%, and 71.2\%, respectively. 
Performance cliffs are in INT(8\textbar3), with accuracies of 23.8\%, 6.1\%, and 35.7\%, respectively.
As a result, the critical nested combination of ResNet is INT(8\textbar4).

As shown in Fig.~\ref{fig:rs_int6_nesting}, for the INT6 nesting ResNet series, compared to FP32 weights models with accuracies of 71.0\%, 75.7\% and 77.7\%, and full-bit models with accuracies of 71.0\%, 75.6\% and 77.5\%, the accuracies of INT(6\textbar5) part-bit models show no degradation.
The INT(6\textbar4) models also have acceptable accuracies of 66.8\%, 67.8\%, and 73.3\%, respectively.
Performance cliffs are in INT(6\textbar3), which accuracies drop to 33.2\%, 7.2\%, and 40.9\%, respectively.
Thus, the critical nested combination is INT(6\textbar4).

\noindent\textbf{DenseNet.} 
For DenseNet series (DenseNet-121/-161/-201) with INT(8\textbar$h$), $h_{\in \{3,4,5,6,7\}}$, as shown in Fig.~\ref{fig:ds_int8_nesting}, compared to FP32 weights and full-bit models with identical accuracies of 76.5\%, 77.2\% and 76.9\%, the part-bit DenseNet-121/-161/-201 of INT(8\textbar7) and INT(8\textbar6), and part-bit DenseNet-121 of INT(8\textbar5) exhibit no performance degradation.
For part-bit DenseNet-161/-201 models, accuracies are 60.0\% and 65.5\% in INT(8\textbar5); When adopting INT(8\textbar4) configuration, accuracies degrade to 0.2\% and 17.1\%, respectively.
The accuracy of the part-bit DenseNet-121 in INT(8\textbar4) is 67.7\%, which degrades to 6.1\% in INT(8\textbar3).
Thus, the critical nested combination of DenseNet-161/-201 is INT(8\textbar5), and for DenseNet-121 is INT(8\textbar4).

The INT6 Nesting in DenseNet is similar, as depicted in Fig.~\ref{fig:ds_int6_nesting}, and the critical nested combination of DenseNet-161/-201 is INT(6\textbar5), and for DenseNet-121 is INT(6\textbar4).

\noindent\textbf{ResNeXt.} The evaluation results of ResNeXt series (ResNeXt-14/-26/-101) are shown in Fig.~\ref{fig:rsx_int8_nesting} and~\ref{fig:rsx_int6_nesting}.
Consistent with the previous analysis, for the INT8 nesting, the critical nested combination of ResNeXt-26/-101 is INT(8\textbar4), while that of ResNeXt-14 is INT(8\textbar5); for the INT6 nesting, the critical nested combination of ResNeXt-26/-101 is INT(6\textbar4), and that of ResNeXt-14 is INT(6\textbar5).

\noindent\textbf{Lightweight Models.} 
The lightweight models have a similar performance in quantization due to the sparse convolutional structure, and we use the same figure for comparison, as shown in Fig~\ref{fig:lw_int8_nesting}.

For \emph{MobileNet}, accuracies of FP32 weights and full-bit models are 73.0\% and 72.9\%, respectively.
The accuracies of INT(8\textbar$h$), $h_{\in \{4,5,6,7\}}$ are 0.2\%, 56.3\%, 69.7\%, and 72.1\%, respectively.
For \emph{MobileNetV2}, the FP32 weights and full-bit models have accuracies of 72.8\% and 72.6\%, respectively.
The part-bit models' accuracies are 3.9\%, 58.7\%, 70.7\%, and 72.2\%, for INT(8\textbar$h$), $h_{\in \{4,5,6,7\}}$, respectively.

For \emph{ShuffleNet}, accuracies of FP32 weights and full-bit models are 64.6\% and 64.8\%, respectively.
The part-bit models' accuracies are 11.8\%, 49.9\%, 63.2\%, and 64.4\%, for INT(8\textbar$h$), $h_{\in \{4,5,6,7\}}$, respectively.
For \emph{ShuffleNetV2}, accuracies of FP32 weights and full-bit models are all 68.3\%.
The accuracies of INT(8\textbar$h$), $h_{\in \{4,5,6,7\}}$ are 21.6\%, 61.1\%, 67.1\%, and 68.2\%, respectively.

For \emph{EfficientNet-B0}, the FP32 weights and full-bit models have accuracies of 56.8\% and 56.4\%, respectively.
The part-bit models' accuracies are 20.4\%, 46.0\%, 54.8\%, and 57.5\%, for INT(8\textbar$h$), $h_{\in \{4,5,6,7\}}$, respectively.

The critical nested combinations of lightweight models are highly similar, which are all in INT(8\textbar5).

\subsubsection{Analysis of Effective/Critical Nested Combinations}\label{ref:analysis_nested_combi}
Based on comprehensive evaluations, for standard CNNs, critical nested combinations are in INT($n$\textbar$\frac{n}{2}$) or INT($n$\textbar$\frac{n}{2}\!+\!1$) for $n\!=\!8$, and INT($n$\textbar$\frac{n}{2}\!+\!1$) or INT($n$\textbar$\frac{n}{2}\!+\!2$) for $n\!=\!6$;
For lightweight CNNs, critical nested combination are in INT($n$\textbar$\frac{n}{2}\!+\!1$) for $n=8$;
For ViTs evaluated in Section~\ref{ref:vit-experiment}, critical nested combinations are in INT($n$\textbar$\frac{n}{2}\!-\!1$) for $n=8$, but with an exception of ViT-B, which is in INT($n$\textbar$\frac{n}{2}$).

\noindent\textbf{Model-Dependence.}
Unfortunately, the effective nested combinations are model dependent, i.e., lightweight CNNs have higher nested bits than standard CNNs, and larger ViTs have even lower nested bits of the critical nested combination.
Thus, we hypothesize that the critical nested combination is related to the model size.

\noindent\textbf{Emerging Patterns.} As the quantitative relationship of ``Critical nested combination vs. Model size'' depicted in Fig.~\ref{fig:robust_nest}, we give a pattern of critical nested combination in Equation~\ref{eq:partbit_pattern}, i.e., the critical nested bit $h$ can be selected with the following rules of thumb: 
\begin{itemize}
    \item[(i)] $h = n/2+1$, for model size in the range $(0, 3\cdot10^1)$; 
    \item[(ii)] $h = n/2$, for model size within $[3\!\cdot\!10^1, 3\!\cdot\!10^2)$; 
    \item[(iii)] $h = n/2-1$, for model size exceeding $3\!\cdot\!10^2$.
\end{itemize}
\noindent Once the critical nested bit is selected, the effective nested combinations can also be determined.
Therefore, we recommend adopting INT($n$\textbar$\frac{n}{2}\!+\!1$) for CNNs and INT($n$\textbar$\frac{n}{2}$) for ViTs to yield the effective NestQuant model in INT8 nesting.

For practically searching for the critical nested combination, we also empirically recommend attempting the INT($n$\textbar$\frac{n}{2}$) initially.
Then, it can search upwards with INT($n$\textbar$\frac{n}{2}\!+\!1$) or downwards with INT($n$\textbar$\frac{n}{2}\!-\!1$) based on the specific performance of INT($n$\textbar$\frac{n}{2}$) differences with the full-bit model.

\subsection{On-Device Resource Measurement}
We chose the ResNet series and lightweight models in effective nested combinations for on-device resource measurement experiments on Raspberry Pi 4B.

\subsubsection{Network Traffic in Transmission}
For the network traffic in transmission, we construct a prototype transmission system between our edge server and IoT device with the transport protocol of TCP/IP socket connection based on the Python interface to evaluate the network traffic.

\begin {figure}[htb!]
\centering
\includegraphics[width=0.85\linewidth]{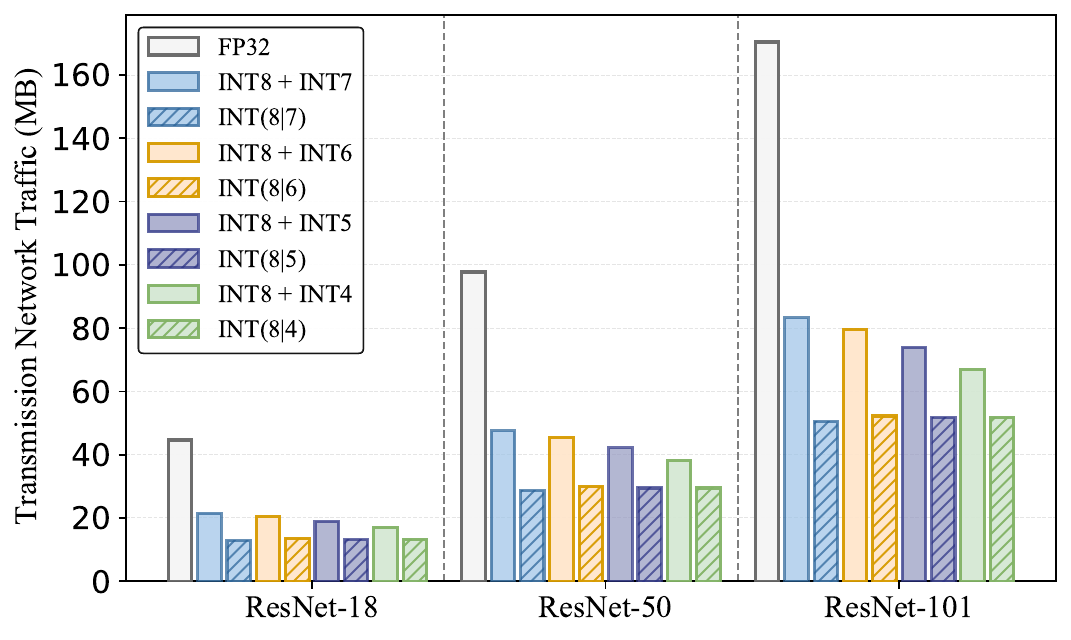}
\caption {Comparison in ResNet series network traffic.} 
\label{fig:transmit_resnet}
\end{figure}

\noindent\textbf{Network Traffic in ResNet Series.} As depicted in Fig.~\ref{fig:transmit_resnet}, the network traffic for transmitting the FP32 ResNet models far exceeds transmitting the other quantized models.
On the deployment of multiple quantized model scenarios, compared to sending diverse bitwidths models, i.e., sending an INT8 model and an INT$h$ model, $h_{\in \{4,5,6,7\}}$, the NestQuant ResNet saves a sizable amount of network traffic.
\begin {figure}[htb!]
\centering
\includegraphics[width=1.\linewidth] {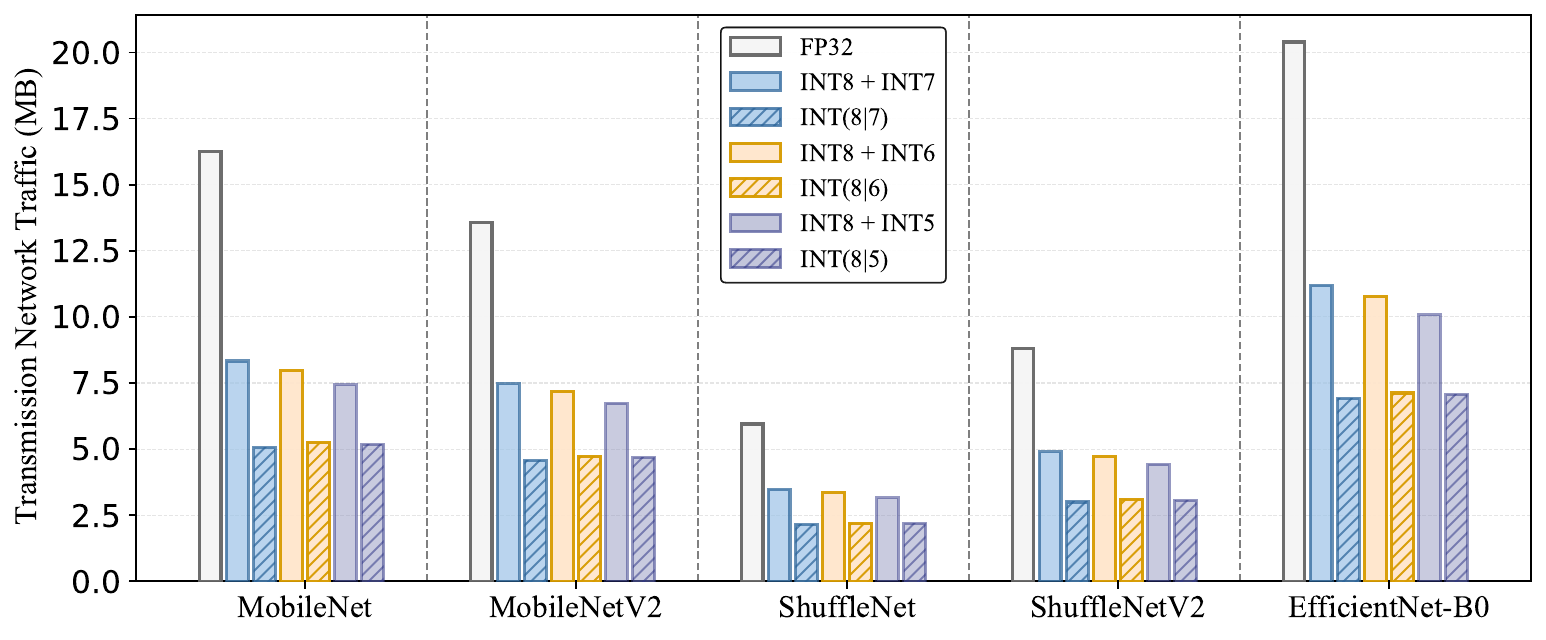}
\caption {Comparison in lightweight models network traffic.} 
\label{fig:transmit_lightweight}
\end{figure}

\noindent\textbf{Network Traffic in Lightweight Models.} As depicted in Fig.~\ref{fig:transmit_lightweight}, the network traffic for transmitting the FP32 lightweight models also far exceeds transmitting the quantized ones.
On the deployment of multiple quantized lightweight models, compared to sending an INT8 model and an INT$h$ model, $h_{\in \{5,6,7\}}$, the NestQuant lightweight models save an amount of network traffic. 
Even for transmitting ShuffleNet series of less than 10MB, NestQuant processed one further reduces network traffic and also enables two model performances at once.

\subsubsection{Model Size in Compression and Deployment}\label{ref:modelsize}
In this paper, the \emph{model size} refers to the disk storage required for model weights.
For the storage of the quantized model, compared to storing diverse bitwidths models, if we only consider the integer weights and ignore the size of FP32 scale $s$ values, the storage reduction in the ideal case can be calculated directly by the bitwidth, as shown in Table~\ref{tab:ideal_int8_storage_redu}.
\begin{table}[htb!]
\centering
\caption{Ideal Nesting Storage Reduction.\label{tab:ideal_int8_storage_redu}}
\resizebox{.35\textwidth}{!}{%
\begin{tabular}{@{\hspace{2mm}}ccc@{\hspace{2mm}}}
\toprule 
\multirow{2}*{\makecell[c]{NestQuant\\Weights}}
& \multirow{2}*{\makecell[c]{Diverse Bitwidths\\Weights}} 
& \multirow{2}*{\makecell[c]{Ideal Storage\\Reduction}}\\
\\ \midrule
INT(8\textbar4) & INT8+INT4 & 25\% \\ 
INT(8\textbar5) & INT8+INT5 & 31\% \\
INT(8\textbar6) & INT8+INT6 & 36\% \\ 
INT(8\textbar7) & INT8+INT7 & 40\% \\  \midrule
INT(6\textbar4) & INT6+INT4 & 30\% \\ 
INT(6\textbar5) & INT6+INT5 & 36\% \\ \bottomrule
\end{tabular}%
}
\end{table}

We perform the practical on-device compression and deployment of NestQuant models by the weights of the packed bits in PyTorch model format \emph{.pth}, and results are shown in Table~\ref{tab:nesting_deploy} and Table~\ref{tab:nesting_deploy_2}.
\begin{table}[ht]
\centering
\caption{Comparison in INT8 Nesting Model Size.\label{tab:nesting_deploy}}
\resizebox{.49\textwidth}{!}{%
\begin{tabular}{@{\hspace{2mm}}lcccccc@{\hspace{2mm}}}
\toprule
\multirow{3}*{\makecell[l]{Model}}&
\multirow{3}*{\makecell[c]{$n$,$h$}}&
\multicolumn{1}{c}{\multirow{1}*{NestQuant}}&
\multicolumn{2}{c}{\multirow{1}*{Diverse Bitwidths}} &
\multicolumn{2}{c}{\multirow{1}*{FP32}}
\\ \cmidrule(lr{0pt}){3-3} \cmidrule(lr{0pt}){4-5} \cmidrule(lr{0pt}){6-7}
&&
\multirow{2}*{\makecell[c]{Model\\Size (MB)}} &
\multirow{2}*{\makecell[c]{Model\\Size (MB)}} &
\multirow{2}*{\makecell[c]{Storage\\Reduction}} &
\multirow{2}*{\makecell[c]{Model\\Size (MB)}} &
\multirow{2}*{\makecell[c]{Storage\\Reduction}}
\\
\\ \midrule
\multirow{4}*{\makecell[l]{ResNet-18}}&{8,4}
& 13.3 & 17.1  & 22.4\% & \multirow{4}*{\makecell[l]{44.7}} & 70.3\% \\ 
&8,5 & 13.3 & 19.0  & 30.0\% & & 70.3\%  \\ 
&8,6 & 13.4  & 20.5  & 34.3\% & & 69.9\% \\
&8,7 & 13.0  & 21.5  & 39.6\% & & 71.0\% \\\midrule
\multirow{4}*{\makecell[l]{ResNet-50}}&{8,4}
& 29.5  & 38.2  & 22.7\%  & \multirow{4}*{\makecell[l]{97.8}} & 69.8\% \\ 
&8,5 & 29.5  & 42.2  & 30.1\% & & 69.8\% \\ 
&8,6 & 29.8  & 45.5  & 34.4\% & & 69.5\% \\
&8,7 & 28.8  & 47.6  & 39.5\% & & 70.5\% \\\midrule
\multirow{4}*{\makecell[l]{ResNet-101}}&{8,4}
& 51.6  & 66.9  & 22.8\% & \multirow{4}*{\makecell[l]{170.5}} & 69.7\%  \\ 
&8,5 & 51.6  & 73.9  & 30.2\% & & 69.7\% \\ 
&8,6 & 52.2  & 79.6  & 34.4\% & & 69.4\% \\
&8,7 & 50.5  & 83.4  & 39.5\% & & 70.4\% \\\midrule
\multirow{3}*{\makecell[l]{MobileNet}}&{8,5}
& 5.2  & 7.4  & 30.4\%  & \multirow{3}*{\makecell[l]{16.3}} & 68.1\% \\ 
&8,6 & 5.2  & 8.0  & 34.3\% & & 67.8\% \\
&8,7 & 5.1  & 8.3  & 39.1\% & & 68.8\% \\\midrule
\multirow{3}*{\makecell[l]{MobileNet\\-V2}}&{8,5}
& 4.7  & 6.7  & 30.6\% & \multirow{3}*{\makecell[l]{13.6}} & 65.5\% \\ 
&8,6 & 4.7  & 7.2  & 34.3\% & & 65.2\% \\
&8,7 & 4.6  & 7.5  & 38.7\% & & 66.2\% \\\midrule
\multirow{3}*{\makecell[l]{ShuffleNet}}&{8,5}
& 2.2  & 3.2  & 31.0\%& \multirow{3}*{\makecell[l]{6.0}} & 63.4\% \\ 
&8,6 & 2.2  & 3.4  & 34.3\% & & 63.0\% \\
&8,7 & 2.1  & 3.5  & 38.5\% & & 64.0\% \\\midrule
\multirow{3}*{\makecell[l]{ShuffleNet\\-V2}}&{8,5}
& 3.1  & 4.4  & 30.8\% & \multirow{3}*{\makecell[l]{8.8}} & 65.3\% \\ 
&8,6 & 3.1  & 4.7  & 34.4\% & & 64.9\% \\
&8,7 & 3.0  & 4.9  & 38.8\% & & 65.9\% \\\midrule
\multirow{3}*{\makecell[l]{EfficientNet\\-B0}}&{8,5}
& 7.1  & 10.1  & 30.1\% & \multirow{3}*{\makecell[l]{20.4}} & 65.4\% \\ 
&8,6 & 7.1  & 10.8  & 33.8\% & & 65.1\% \\
&8,7 & 6.9  & 11.2  & 38.3\% & & 66.1\% \\
\bottomrule
\end{tabular}%
}
\end{table}

\noindent\textbf{INT8 Nesting Model Size.} 
For ResNet series in INT(8\textbar$h$), $h_{\in \{4,5,6,7\}}$ configurations, compared to the model size of storing diverse bitwidths models, 
ResNet-18 saves 22.4\%, 30.0\%, 34.3\%, and 39.6\%; 
ResNet-50 can reduce 22.7\%, 30.1\%, 34.4\%, and 39.5\%; 
ResNet-101 can save 22.8\%, 30.2\%, 34.4\%, and 39.5\%, respectively.

For lightweight models in INT(8\textbar$h$), $h_{\in \{5,6,7\}}$ configurations, compared to the model size of storing diverse bitwidths models, 
MobileNet saves 30.4\%, 34.3\%, and 39.1\%; 
MobileNetV2 can reduce 30.6\%, 34.3\%, and 38.7\%; 
ShuffleNet can save 31.0\%, 34.3\%, and 38.5\%; 
ShuffleNetV2 reduces 30.8\%, 34.4\%, and 38.8\%; 
EfficientNet-B0 can save 30.1\%, 33.8\%, and 38.3\%, respectively.

\begin{table}[ht]
\centering
\caption{Comparison in INT6 Nesting Model Size.\label{tab:nesting_deploy_2}}
\resizebox{.49\textwidth}{!}{%
\begin{tabular}{@{\hspace{2mm}}lcccccc@{\hspace{2mm}}}
\toprule
\multirow{3}*{\makecell[l]{Model}}&
\multirow{3}*{\makecell[c]{$n$,$h$}}&
\multicolumn{1}{c}{\multirow{1}*{NestQuant}}&
\multicolumn{2}{c}{\multirow{1}*{Diverse Bitwidths}} &
\multicolumn{2}{c}{\multirow{1}*{FP32}}
\\ \cmidrule(lr{0pt}){3-3} \cmidrule(lr{0pt}){4-5} \cmidrule(lr{0pt}){6-7}
&&
\multirow{2}*{\makecell[c]{Model\\Size (MB)}} &
\multirow{2}*{\makecell[c]{Model\\Size (MB)}} &
\multirow{2}*{\makecell[c]{Storage\\Reduction}} &
\multirow{2}*{\makecell[c]{Model\\Size (MB)}} &
\multirow{2}*{\makecell[c]{Storage\\Reduction}}
\\
\\ \midrule
\multirow{2}*{\makecell[l]{ResNet-18}}& {6,4} & 10.1 & 14.9 & 32.2\% & \multirow{2}*{\makecell[l]{44.7}} & 77.4\% \\ 
&{6,5} & 10.5 & 16.7 & 37.4\% && 76.5\%\\ \midrule
\multirow{2}*{\makecell[l]{ResNet-50}}& {6,4} & 22.5 & 33.3 & 32.3\% & \multirow{2}*{\makecell[l]{97.8}} & 76.9\% \\ 
&{6,5} & 23.4 & 37.3 & 37.3\% && 76.1\%\\ \midrule
\multirow{2}*{\makecell[l]{ResNet-101}}& {6,4} & 39.5 & 58.4 & 32.3\% & \multirow{2}*{\makecell[l]{170.5}} & 76.8\% \\ 
&{6,5} & 41.0 & 65.5 & 37.3\% && 75.9\%\\ \bottomrule
\end{tabular}%
}
\end{table}

\begin{table*}[htbp!]
\centering
\caption{Numerical Computation of Switching Overheads and Memory Usage.\label{tab:switching_cost}}
\resizebox{.98\textwidth}{!}{%
\begin{tabular}{@{\hspace{2mm}}lcccccccccccccc@{\hspace{2mm}}}
\toprule
\multirow{4}*{\makecell[l]{Model}}&
\multirow{4}*{\makecell[c]{$n$,$h$}}&
\multicolumn{6}{c}{\multirow{1}*{Part-Bit Model Upgrade Switching Overhead (MB)}}&
\multicolumn{6}{c}{\multirow{1}*{Full-Bit Model Downgrade Switching Overhead (MB)}}
&\multirow{1}*{\makecell[c]{FP32}}
\\ \cmidrule(lr{0pt}){3-8} \cmidrule(lr{0pt}){9-14} \cmidrule(lr{0pt}){15-15}
&&
\multirow{2}*{\makecell[c]{Memory\\Usage}}&
\multicolumn{2}{c}{\multirow{1}*{NestQuant}}&
\multicolumn{2}{c}{\multirow{1}*{Diverse Bitwidths}}&
\multirow{2}*{\makecell[c]{Reduced\\Overhead}}&
\multirow{2}*{\makecell[c]{Memory\\Usage}}&
\multicolumn{2}{c}{\multirow{1}*{NestQuant}}&
\multicolumn{2}{c}{\multirow{1}*{Diverse Bitwidths}}&
\multirow{2}*{\makecell[c]{Reduced\\Overhead}}&
\multirow{2}*{\makecell[c]{Memory\\Usage}}
\\ 
\cmidrule(lr{0pt}){4-5}\cmidrule(lr{0pt}){6-7} \cmidrule(lr{0pt}){10-11}\cmidrule(lr{0pt}){12-13}
&&&
\multirow{1}*{\makecell[c]{Page-in}}&
\multirow{1}*{\makecell[c]{Page-out}}&
\multirow{1}*{\makecell[c]{Page-in}}&
\multirow{1}*{\makecell[c]{Page-out}}&
&&
\multirow{1}*{\makecell[c]{Page-in}}&
\multirow{1}*{\makecell[c]{Page-out}}&
\multirow{1}*{\makecell[c]{Page-in}}&
\multirow{1}*{\makecell[c]{Page-out}}&
\\
\midrule
\multirow{6}*{\makecell[l]{ResNet-18}}
& 8,4 & 43.1 & 7.4 & 0 & 11.3 & 5.8 & 56.9\% & 86.2 & 0 & 7.4 & 5.8 & 11.3 & 56.9\% 
& \multirow{6}*{\makecell[c]{147.2}}\\ 
& 8,5 & 53.9 & 5.9 & 0 & 11.3 & 7.6 & 68.9\% & 86.2 & 0 & 5.9 & 7.6 & 11.3 & 68.9\% \\ 
& 8,6 & 64.7 & 4.5 & 0 & 11.3 & 9.1 & 78.1\% & 86.2 & 0 & 4.5 & 9.1 & 11.3 & 78.1\% \\
& 8,7 & 75.5 & 2.9 & 0 & 11.3 & 10.1 & 86.6\% & 86.2 & 0 & 2.9 & 10.1 & 11.3 & 86.6\% \\
& 6,4 & 43.1 & 5.1 & 0 & 9.1 & 5.8 & 66.1\% & 64.7 & 0 & 5.1 & 5.8 & 9.1 & 66.1\% \\
& 6,5 & 53.9 & 3.5 & 0 & 9.1 & 7.6 & 79.1\% & 64.7 & 0 & 3.5 & 7.6 & 9.1  & 79.1\% \\
\midrule
\multirow{6}*{\makecell[l]{ResNet-50}}
& 8,4 & 56.2 & 16.4 & 0 & 25.2 & 13.0 & 57.1\% & 112.3 & 0 & 16.4 & 13.0 & 25.2 & 57.1\% 
& \multirow{6}*{\makecell[c]{283.3}}\\ 
& 8,5 & 70.2 & 13.1 & 0 & 25.2 & 17.1 & 69.0\% & 112.3 & 0 & 13.1 & 17.1 & 25.2 & 69.0\% \\ 
& 8,6 & 84.2 & 9.9 & 0 & 25.2 & 20.3 & 78.1\% & 112.3 & 0 & 9.9 & 20.3 & 25.2 & 78.1\% \\
& 8,7 & 98.3 & 6.4 & 0 & 25.2 & 22.5 & 86.6\% & 112.3 & 0 & 6.4 & 22.5 & 25.2 & 86.6\% \\
& 6,4 & 56.2 & 11.3 & 0 & 20.3 & 13.0 & 66.2\% & 84.2 & 0 & 11.3 & 13.0 & 20.3 & 66.2\%\\
& 6,5 & 70.2 & 7.8 & 0 & 20.3 & 17.1 & 79.1\% & 84.2 & 0 & 7.8 & 17.1 & 20.3 & 79.1\% \\
\midrule
\multirow{6}*{\makecell[l]{ResNet-101}}
& 8,4 & 72.0 & 28.7 & 0 & 44.0 & 22.8 & 57.1\% & 144.1 & 0 & 28.7 & 22.8 & 44.0 & 57.1\%
& \multirow{6}*{\makecell[c]{402.9}}\\  
& 8,5 & 90.0 & 23.0 & 0 & 44.0 & 29.9 & 69.0\% & 144.1 & 0 & 23.0 & 29.9 & 44.0 & 69.0\% \\ 
& 8,6 & 108.1 & 17.4 & 0 & 44.0 & 35.6 & 78.1\% & 144.1 & 0 & 17.4 & 35.6 & 44.0 & 78.1\% \\
& 8,7 & 126.1 & 11.2 & 0 & 44.0 & 39.3 & 86.6\% & 144.1 & 0 & 11.2 & 39.3 & 44.0 & 86.6\% \\
& 6,4 & 72.0 & 19.8 & 0 & 35.6 & 22.8 & 66.2\% & 108.1 & 0 & 19.8 & 22.8 & 35.6 & 66.2\% \\
& 6,5 & 90.0 & 13.7 & 0 & 35.6 & 29.9 & 79.1\% & 108.1 & 0 & 13.7 & 29.9 & 35.6 & 79.1\% \\
\midrule
\multirow{3}*{\makecell[l]{MobileNet}}
& 8,5 & 42.0 & 2.3 & 0 & 4.4 & 3.1 & 69.1\% & 67.1 & 0 & 2.3 & 3.1 & 4.4 & 69.1\%
& \multirow{3}*{\makecell[c]{86.1}}\\  
& 8,6 & 50.3 & 1.8 & 0 & 4.4 & 3.6 & 78.1\% & 67.1 & 0 & 1.8 & 3.6 & 4.4 & 78.1\% \\
& 8,7 & 58.7 & 1.1 & 0 & 4.4 & 3.9 & 86.5\% & 67.1 & 0 & 1.1 & 3.9 & 4.4 & 86.5\% \\
\midrule
\multirow{3}*{\makecell[l]{MobileNet\\-V2}}
& 8,5 & 42.9 & 2.1 & 0 & 3.9 & 2.8 & 69.1\% & 68.6 & 0 & 2.1 & 2.8 & 3.9 & 69.1\%
& \multirow{3}*{\makecell[c]{93.1}}\\
& 8,6 & 51.5 & 1.6 & 0 & 3.9 & 3.3 & 78.1\% & 68.6 & 0 & 1.6 & 3.3 & 3.9 & 78.1\% \\
& 8,7 & 60.0 & 1.0 & 0 & 3.9 & 3.6 & 86.4\% & 68.6 & 0 & 1.0 & 3.6 & 3.9 & 86.4\% \\
\midrule
\multirow{3}*{\makecell[l]{ShuffleNet}}
& 8,5 & 40.7 & 1.0 & 0 & 1.8 & 1.3 & 69.3\% & 65.1 & 0 & 1.0 & 1.3 & 1.8 & 69.3\%
& \multirow{3}*{\makecell[c]{71.4}}\\
& 8,6 & 48.8 & 0.7 & 0 & 1.8 & 1.5 & 78.1\% & 65.1 & 0 & 0.7 & 1.5 & 1.8 & 78.1\% \\
& 8,7 & 57.0 & 0.5 & 0 & 1.8 & 1.7 & 86.3\% & 65.1 & 0 & 0.5 & 1.7 & 1.8 & 86.3\% \\
\midrule
\multirow{3}*{\makecell[l]{ShuffleNet\\-V2}}
& 8,5 & 41.9 & 1.4 & 0 & 2.6  & 1.9 & 69.2\% & 67.0 & 0 & 1.4 & 1.9 & 2.6 & 69.2\%
& \multirow{3}*{\makecell[c]{77.4}}\\
& 8,6 & 50.3 & 1.0 & 0 & 2.6  & 2.1 & 78.1\% & 67.0 & 0 & 1.0 & 2.1 & 2.6 & 78.1\%  \\
& 8,7 & 58.6 & 0.7 & 0 & 2.6  & 2.3 & 86.4\% & 67.0 & 0 & 0.7 & 2.3 & 2.6 & 86.4\% \\
\midrule
\multirow{3}*{\makecell[l]{EfficientNet\\-B0}}
& 8,5 & 65.9 & 3.1 & 0 & 5.9 & 4.2 & 69.0\% & 105.4 & 0 & 3.1 & 4.2 & 5.9 & 69.0\%
& \multirow{3}*{\makecell[c]{126.5}}\\
& 8,6 & 79.0 & 2.4 & 0 & 5.9 & 4.9 & 78.0\% & 105.4 & 0 & 2.4 & 4.9 & 5.9 & 78.0\% \\
& 8,7 & 92.2 & 1.5 & 0 & 5.9 & 4.2 & 86.3\% & 105.4 & 0 & 1.5 & 5.3 & 5.9 & 86.3\%  \\
\bottomrule
\end{tabular}%
}
\end{table*}

As a result, INT8 nesting models can approximate the ideal storage reductions 25\%, 31\%, 36\% and 40\% in Table~\ref{tab:ideal_int8_storage_redu}.

\noindent\textbf{INT6 Nesting Model Size.} 
The results of INT6 nesting deployment are also close to the ideal storage reductions in Table~\ref{tab:ideal_int8_storage_redu}.
In INT(6\textbar4) and INT(6\textbar5) configurations, compared to the model size of storing diverse bitwidths models, 
ResNet-18 can save 32.2\% and 37.4\%; 
ResNet-50 saves 32.3\% and 37.3\%; 
ResNet-101 can reduce 32.3\% and 37.3\%, respectively.

Compared to the FP32 models, the NestQuant models in INT6/INT8 nesting deployment bring the same significant storage reductions due to the gain in quantization.
In summary, the NestQuant model is storage resource-friendly for on-device compression and deployment.

\begin{table}[htb!]
\centering
\caption{Evaluation in INT8 Nesting ViTs.\label{tab:nestquant_vit}}
\resizebox{.46\textwidth}{!}{%
\begin{tabular}{@{\hspace{2mm}}llcr@{\textbar}lcr@{\textbar}l@{\hspace{2mm}}}
\toprule
\multirow{2}*{\makecell[l]{}}&
\multirow{2}*{\makecell[l]{A-bit}}& 
\multirow{2}*{\makecell[c]{W-bit}}&
\multicolumn{2}{c}{\makecell[c]{Top-1 Acc. (\%)}}&
\multirow{2}*{\makecell[c]{Model\\Size (MB)}}&
\multicolumn{2}{c}{\makecell[c]{Mem. Usage (MB)}}\\ \cmidrule{4-5} \cmidrule{7-8}
&&& \multirow{1}*{\makecell[c]{Part-Bit}} 
& \multirow{1}*{\makecell[c]{Full-Bit}}& &\multirow{1}*{\makecell[c]{Part-Bit}} 
& \multirow{1}*{\makecell[c]{Full-Bit}}\\
\midrule
\multirow{8}*{\rotatebox{90}{\makecell[l]{DeiT-B}}}& FP32
& FP32 & \makecell[c]{-} & 81.7& 330.3 & \makecell[c]{-} & 732.4\\ \cmidrule{2-8}
&\multirow{7}*{\makecell[l]{INT8}} & INT8 & \makecell[c]{-} & 80.5 & 84.2 & \makecell[c]{-} & 251.9\\
&& INT(8\textbar7) & 80.4 & 80.5 & 96.3 & 220.4 & 251.9\\
&& INT(8\textbar6) & 80.3 & 80.5 & 99.7 & 188.9 & 251.9\\ 
&& INT(8\textbar5) & 80.2 & 80.5 & 98.6 & 157.4 & 251.9\\
&& INT(8\textbar4) & 79.6 & 80.5 & 98.6 & 126.0 & 251.9\\ 
&& INT(8\textbar3) & 74.6 & 80.5 & 99.7 & 94.5 & 251.9\\ 
&& INT(8\textbar2) & 2.3 & 80.5  & - & \makecell[c]{-} &\makecell[c]{-} \\
\midrule
\multirow{8}*{\rotatebox{90}{\makecell[l]{Swin-B}}}& FP32
& FP32 & \makecell[c]{-} & 84.7 & 336.5 & \makecell[c]{-} & 806.5 \\\cmidrule{2-8}
&\multirow{7}*{\makecell[l]{INT8}} & INT8 & \makecell[c]{-} & 84.0 & 87.1 & \makecell[c]{-} & 349.2 \\
&& INT(8\textbar7) & 84.0 & 84.0 & 99.6 & 305.6 & 349.2 \\
&& INT(8\textbar6) & 84.0 & 84.0 & 103.1 & 261.9 & 349.2 \\ 
&& INT(8\textbar5) & 84.0 & 84.0 & 101.9 & 218.3 & 349.2 \\
&& INT(8\textbar4) & 83.6 & 84.0 & 101.9 & 174.6 & 349.2 \\ 
&& INT(8\textbar3) & 79.4 & 84.0 & 103.1 & 131.0 & 349.2 \\ 
&& INT(8\textbar2) & 0.7 & 84.0  & - & \makecell[c]{-} & \makecell[c]{-}\\
\midrule
\multirow{7}*{\rotatebox{90}{\makecell[l]{ViT-B}}}& FP32& FP32 & \makecell[c]{-} & 78.0 & 330.3 & \makecell[c]{-} & 732.4 \\\cmidrule{2-8}
&\multirow{6}*{\makecell[l]{INT8}} & INT8 & \makecell[c]{-} & 75.7 & 84.2 & \makecell[c]{-} & 252.2 \\
&& INT(8\textbar7) & 75.5 & 75.7 & 96.3 & 220.6 & 252.2 \\
&& INT(8\textbar6) & 75.3 & 75.7 & 99.7 & 189.1 & 252.2 \\ 
&& INT(8\textbar5) & 74.9 & 75.7 & 98.6 & 157.6 & 252.2 \\
&& INT(8\textbar4) & 69.9 & 75.7 & 98.6 & 126.1 & 252.2 \\ 
&& INT(8\textbar3) & 24.3 & 75.7 & - & \makecell[c]{-} & \makecell[c]{-}\\ 
\midrule
\multirow{8}*{\rotatebox{90}{\makecell[l]{Swin-L}}}& FP32
& FP32 & \makecell[c]{-} & 85.8& 751.4 & \makecell[c]{-} & 1608.1\\\cmidrule{2-8}
&\multirow{7}*{\makecell[l]{INT8}} & INT8 & \makecell[c]{-} & 84.5 & 191.5 & \makecell[c]{-} & 571.7 \\
&& INT(8\textbar7) & 84.5 & 84.5 & 218.9 & 500.3 & 571.7 \\
&& INT(8\textbar6) & 84.5 & 84.5 & 226.8 & 428.8 & 571.7 \\ 
&& INT(8\textbar5) & 84.5 & 84.5 & 224.1 & 357.3 & 571.7 \\
&& INT(8\textbar4) & 84.3 & 84.5 & 224.1 & 285.9 & 571.7 \\ 
&& INT(8\textbar3) & 82.1 & 84.5 & 226.8 & 214.4 & 571.7 \\ 
&& INT(8\textbar2) & 17.6 & 84.5  & - & \makecell[c]{-} & \makecell[c]{-} \\
\midrule
\multirow{8}*{\rotatebox{90}{\makecell[l]{ViT-L}}}&FP32
& FP32 & \makecell[c]{-} & 84.4& 1161.0 & \makecell[c]{-} & 2445.2\\\cmidrule{2-8} 
&\multirow{7}*{\makecell[l]{INT8}} & INT8 & \makecell[c]{-} & 82.4 & 293.8 & \makecell[c]{-} & 694.4 \\
&& INT(8\textbar7) & 82.5 & 82.4 & 335.9 & 607.6 & 694.4 \\
&& INT(8\textbar6) & 82.5 & 82.4 & 348.0 & 520.8 & 694.4 \\ 
&& INT(8\textbar5) & 82.3 & 82.4 & 343.9 & 434.0 & 694.4 \\
&& INT(8\textbar4) & 81.6 & 82.4 & 343.9 & 347.2 & 694.4 \\ 
&& INT(8\textbar3) & 77.7 & 82.4 & 348.0 & 260.4 & 694.4 \\
&& INT(8\textbar2) & 1.4 & 82.4 & - & \makecell[c]{-} & \makecell[c]{-}\\
\bottomrule
\end{tabular}%
}
\end{table}

\begin{table*}[ht]
\centering
\caption{Comparison of Mixed/Dynamic Precision Quantization in ResNet-18/-50 (The asterisk (*) denotes data not reported in the original paper; The ``HW'', i.e., special hardware).\label{tab:comparison_dynamic}}
\resizebox{.95\textwidth}{!}{%
\begin{tabular}{@{\hspace{2mm}}lllccccccc@{\hspace{2mm}}}
\toprule
\multirow{2}*{\makecell[l]{}}& 
\multirow{2}*{\makecell[l]{Tech.}}&
\multirow{2}*{\makecell[l]{Method}}&
\multirow{2}*{\makecell[l]{A-bit}}& 
\multirow{2}*{\makecell[c]{W-bit}}&
\multirow{2}*{\makecell[c]{ImageNet-1K Top-1 Acc. (\%)}}& 
\multicolumn{3}{c}{\makecell[c]{Require}}&
\multirow{2}*{\makecell[l]{Model Size}}
\\ \cmidrule{7-9}
&&&& & &Train& Data & HW &\\
\midrule
\multirow{10}*{\rotatebox{90}{ResNet-18}} & \multirow{1}*{-} & Pretrained & FP32 & FP32 & 71.5 & - & - & -&  FP32 (44.7MB) \\ \cmidrule{2-10}
&\multirow{2}*{QAT} & AnyPrecision~\cite{yu2021any}& INT[8,4,2,1] & INT[8,4,2,1] & 68.0/68.0/64.2/54.6& \CheckmarkBold & \CheckmarkBold & \XSolidBrush& FP32\\ 
&& EQ-Net~\cite{xu2023eq}& INT[8,7,6,5,4,3,2] & INT[8,7,6,5,4,3,2] & 70.7/70.7/70.8/70.6/70.3/69.3/65.9 &\CheckmarkBold & \CheckmarkBold & \XSolidBrush& FP32\\  \cmidrule{2-10}
&\multirow{1}*{MP} & SPARK~\cite{liu24spark}& INT4 MP & INT4 MP & 69.7 & \XSolidBrush & \XSolidBrush & \CheckmarkBold &  -* \\ \cmidrule{2-10}
&\multirow{5}*{PTQ} & BRECQ~\cite{li2021brecq}& INT8 & INT8 & 71.0 & \XSolidBrush & \CheckmarkBold & \XSolidBrush& INT8 (11.3MB)\\
&& OBQ~\cite{frantar2022optimal}& INT8 & INT8 & 69.7 & \XSolidBrush & \CheckmarkBold & \XSolidBrush&   INT8\\
&& SQuant~\cite{guo2022squant}& INT8 & INT8 & 71.4 & \XSolidBrush & \XSolidBrush & \XSolidBrush&   INT8\\
&& Diverse Bitwidths& INT8+INT4 & INT8+INT4 & 71.4/64.2 & \XSolidBrush  & \XSolidBrush & \XSolidBrush&  INT8+INT4 (17.1MB)\\
&& \cellcolor{gray!20}\textbf{NestQuant (Ours)}& \cellcolor{gray!20}INT8 & \cellcolor{gray!20}INT(8\textbar4) & \cellcolor{gray!20}71.4/64.1 & \cellcolor{gray!20}\XSolidBrush  & \cellcolor{gray!20}\XSolidBrush & \cellcolor{gray!20}\XSolidBrush & \cellcolor{gray!20}INT4+INT5 (13.3MB)\\ \hline

\multirow{9}*{\rotatebox{90}{ResNet-50}} & \multirow{1}*{-} & Pretrained & FP32 & FP32 & 77.7 & - & - & - &  FP32 (97.8MB) \\ \cmidrule{2-10}
&\multirow{3}*{QAT} & AdaBits~\cite{jin2020adabits}& INT[4,3,2] & INT[4,3,2] & 76.1/75.8/73.2 & \CheckmarkBold &  \CheckmarkBold& \XSolidBrush &  FP32\\ 
&& AnyPrecision~\cite{yu2021any}&INT[8,4,2,1] & INT[8,4,2,1] &  74.9/74.8/73.2/63.2& \CheckmarkBold & \CheckmarkBold& \XSolidBrush & FP32\\ 
&& EQ-Net~\cite{xu2023eq}& INT[8,7,6,5,4,3,2] & INT[8,7,6,5,4,3,2] & 75.4/75.6/75.6/75.5/75.3/74.7/72.6 &\CheckmarkBold & \CheckmarkBold& \XSolidBrush & FP32\\  \cmidrule{2-10}
&\multirow{1}*{MP} & SPARK~\cite{liu24spark}& INT4 MP & INT4 MP & 76.1 & \XSolidBrush & \XSolidBrush & \CheckmarkBold &  -* \\ \cmidrule{2-10}
&\multirow{3}*{PTQ} & SQuant~\cite{guo2022squant}& INT8 & INT8 & 77.7 & \XSolidBrush & \XSolidBrush& \XSolidBrush &  INT8 (25.2MB)\\
&& Diverse Bitwidths& INT8+INT4 & INT8+INT4 & 77.7/68.6 & \XSolidBrush  & \XSolidBrush& \XSolidBrush &  INT8+INT4 (38.2MB)\\
&& \cellcolor{gray!20}\textbf{NestQuant (Ours)}& \cellcolor{gray!20}INT8 & \cellcolor{gray!20}INT(8\textbar4) & \cellcolor{gray!20}77.7/66.3 & \cellcolor{gray!20}\XSolidBrush  & \cellcolor{gray!20}\XSolidBrush & \cellcolor{gray!20}\XSolidBrush& \cellcolor{gray!20}INT4+INT5 (29.5MB)\\ 
\bottomrule
\end{tabular}%
}
\end{table*}

\subsubsection{Numerical Computation of Switching Overheads}
The switching overheads between memory and storage can have a significant impact on the quality of service.
The overheads of model switching can be numerically calculated by disk size, and the premise of numerical computations is that the size of packed-bit weights indeed approximates the ideal quantized model size.
We have verified this premise in Section~\ref{ref:modelsize}.
For the numerical computations, assuming the $P^{(\mathrm{in})}$ is the page-in overhead, and the $D$ is the packed-bit model disk size.
The primitive page-in overheads of $\bm{w}_\mathrm{high}$ can be calculated by:
$P^{(\mathrm{in})}_\mathrm{high} \!=\! D_\mathrm{high} \!\approx\! \frac{h}{h + l + 1} \!\cdot\! D$; 
the upgrading page-in overheads of $\bm{w}_\mathrm{low}$ can be calculated by:
$P^{(\mathrm{in})}_\mathrm{low} \!=\! D_\mathrm{low}\!\approx\! \frac{l + 1}{h + l + 1} \!\cdot\! D$.
For the downgrading page-out overhead, its value should be the same as the upgrade page-in overheads of $P^{(\mathrm{in})}_\mathrm{low}$ in theory.
For diverse bitwidths models, numerical page-in and page-out overheads correspond to the two exchanging packed-bit INT model disk sizes, respectively.
We numerically calculate the switching overhead of page-in and page-out in memory when upgrading and downgrading the NestQuant model, and also compared the overhead required by diverse bitwidth models to switch models in the same situation, as depicted in Table~\ref{tab:switching_cost}.

We also report the CPU memory usages $U$ of both FP32 ($U_\mathrm{fp32}$) and INT8 ($U_\mathrm{int8}$) models during inference with a batch size of one in Table~\ref{tab:switching_cost} and Table~\ref{tab:nestquant_vit}.
For example, the memory usage of FP32 ViT-L model with 1.13 GB model size is $U_\mathrm{fp32}=2445.2$MB as shown in Table~\ref{tab:nestquant_vit}.
While the memory usages of other $k$-bit models are estimated based on the memory usage of INT8 model as $U_{\mathrm{int}k}= \frac{k}{8}\cdot U_\mathrm{int8}$ for $k\ {\in \{3,4,...,7\}}$.

Theoretically, the NestQuant model has a zero page-out overhead for upgrades and a zero page-in overhead for downgrades because it is nested. 
For a comparison, diverse bitwidths models have to pay extra switching overheads in previous cases.
As the page-in overhead of upgrading corresponds to the page-out overhead of downgrading, we only need to discuss one of these reduced overheads.

\noindent\textbf{Reduced Overheads in ResNet Series.} 
Compared to switching overheads of diverse bitwidths models in INT(8\textbar$h$), $h_{\in \{4,5,6,7\}}$, and INT(6\textbar$h$), $h_{\in \{4,5\}}$,
ResNet-18 reduces 56.9\%, 68.9\%, 78.1\%, 86.6\%, 66.1\%, and 79.1\%; 
ResNet-50 can save 57.1\%, 69.0\%, 78.1\%, 86.6\%, 66.2\%, and 79.1\%;
ResNet-101 can reduce 57.1\%, 68.9\%, 78.1\%, 86.6\%, 66.1\%, and 79.1\%, respectively.

\noindent\textbf{Reduced Overheads in Lightweight Models.} Compared to switching overheads of diverse bitwidths models in INT(8\textbar$h$), $h_{\in \{5,6,7\}}$,
MobileNet saves 69.1\%, 78.1\%, and 86.5\%; 
MobileNetV2 can reduce 69.1\%, 78.1\%, and 86.4\%; 
ShuffleNet can save 69.3\%, 78.1\%, and 86.3\%; 
ShuffleNetV2 reduces 69.2\%, 78.1\%, and 86.4\%; 
EfficientNet-B0 can save 69.0\%, 78.0\%, and 86.3\%, respectively.

In summary, the capability of model switching can satisfy the requirement of resource adaptation, while the NestQuant models reduce switching overheads significantly compared to the diverse bitwidths models, mitigating the memory page-in/-out constraints of the on-device services.

\subsection{Evaluation in Vision Transformer Models}\label{ref:vit-experiment} 
The larger ViT models via nesting quantization also have comparable performance in ImageNet-1K, as shown in Table~\ref{tab:nestquant_vit}.
For the \emph{Swin-B/-L}, \emph{DeiT-B}, and \emph{ViT-L}, the critical nested combinations are all in INT(8\textbar3).
For the \emph{ViT-B}, the critical nested combination is in INT(8\textbar4);
Besides, compared to NestQuant CNNs, the NestQuant ViTs have a larger model size and have more capabilities of lower precision quantization, so the critical nested combination is in a more lower bit of INT($n$\textbar$\frac{n}{2}\!-\!1$) for $n=8$.

\subsection{Comparison of Mixed/Dynamic Precision Methods.} 
We consider the mixed precision (MP) method \emph{SPARK}~\cite{liu24spark}, which requires special hardware support, and dynamic precision quantization methods \emph{AdaBits}~\cite{jin2020adabits}, \emph{AnyPrecision DNN}~\cite{yu2021any}, and \emph{EQ-Net}~\cite{xu2023eq}, which are QAT methods requiring resource-intensive training and full training datasets, while our proposed NestQuant focuses on PTQ.

As shown in Table~\ref{tab:comparison_dynamic}, the dynamic precision methods for DNN can get better model performances but consume a lot of resources (i.e., computation, data) for training a switchable quantized DNN model.
These methods also need to store the FP32 disk size model for multiple bitwidths switching.

For MP methods, specifically, the SPARK\cite{liu24spark} operates by element-wise separation of weight values into low-order parts within $[0,7]$ and high-order parts within $[8,255]$ based on value ranges, so the values of low-order parts can be encoded to 4-bit, and the values of high-order parts are encoded to 8-bit, requiring special hardware to handle the mixed 4-/8-bit inference.
The SPARK can retain the highest accuracy with approximating INT4 occupations (i.e., INT4 MP) without training or data, but needs hardware support, whereas NestQuant is more hardware-friendly in the absence of special hardware.
The NestQuant method considers the nesting in PTQ and has the advantage of quickly optimizing a two-precision switching quantized DNN model without data, training, or special hardware.
Therefore, from this perspective, the NestQuant is more deployment-friendly for on-device DNNs.

\section{Conclusions and Future work} \label{ref:5-conclusion}
This paper introduces a resource-friendly post-training integer-nesting quantization, NestQuant, to provide a switchable on-device quantized DNN model for adapting multi-scenarios.
The NestQuant contains an integer weight decomposition for splitting weights into lower-bit and higher-bit weights, and nests the higher-bit weights into original weights, so it can switch between two instances: full-bit and part-bit models.
We explored the effective nested combinations for the part-bit model and performance compensation for the full-bit model.
Experimental results demonstrate that the performance of the part-bit model does not degrade a lot compared with the full-bit model or even the FP32 model in effective nested combinations.
This provides a solid performance basis for model switching between different bitwidths.
On IoT device resource measurements, the NestQuant model outperforms diverse bitwidths PTQ models in terms of network traffic, model size, and switching overheads.

In the future, we will attempt to migrate such nesting quantized model switching techniques into the compression of storage-sensitive on-device large models.
We will also try to explore the adaptive nesting selection scheme for finding the optimal NestQuant combinations automatically.


\ifCLASSOPTIONcaptionsoff
  \newpage
\fi

\bibliographystyle{IEEEtran}
\bibliography{IEEEtran/IEEEabrv,IEEEtran/main}

\begin{thebibliography}{10}
\providecommand{\url}[1]{#1}
\csname url@samestyle\endcsname
\providecommand{\newblock}{\relax}
\providecommand{\bibinfo}[2]{#2}
\providecommand{\BIBentrySTDinterwordspacing}{\spaceskip=0pt\relax}
\providecommand{\BIBentryALTinterwordstretchfactor}{4}
\providecommand{\BIBentryALTinterwordspacing}{\spaceskip=\fontdimen2\font plus
\BIBentryALTinterwordstretchfactor\fontdimen3\font minus \fontdimen4\font\relax}
\providecommand{\BIBforeignlanguage}[2]{{%
\expandafter\ifx\csname l@#1\endcsname\relax
\typeout{** WARNING: IEEEtran.bst: No hyphenation pattern has been}%
\typeout{** loaded for the language `#1'. Using the pattern for}%
\typeout{** the default language instead.}%
\else
\language=\csname l@#1\endcsname
\fi
#2}}
\providecommand{\BIBdecl}{\relax}
\BIBdecl

\bibitem{xu2020deepwear}
M.~Xu, F.~Qian, M.~Zhu, F.~Huang, S.~Pushp, and X.~Liu, ``Deepwear: Adaptive local offloading for on-wearable deep learning,'' \emph{IEEE Transactions on Mobile Computing}, vol.~19, no.~2, pp. 314--330, 2020.

\bibitem{Liu@ANew}
C.~Liu, Y.~Cao, Y.~Luo, G.~Chen, V.~Vokkarane, M.~Yunsheng, S.~Chen, and P.~Hou, ``A new deep learning-based food recognition system for dietary assessment on an edge computing service infrastructure,'' \emph{IEEE Transactions on Services Computing}, vol.~11, no.~2, pp. 249--261, 2018.

\bibitem{xu2019Afirst}
M.~Xu, J.~Liu, Y.~Liu, F.~X. Lin, Y.~Liu, and X.~Liu, ``A first look at deep learning apps on smartphones,'' in \emph{Proceedings of the World Wide Web Conference}, 2019, p. 2125–2136.

\bibitem{zhang2024Acomprehensive}
Q.~Zhang, X.~Che, Y.~Chen, X.~Ma, M.~Xu, S.~Dustdar, X.~Liu, and S.~Wang, ``A comprehensive deep learning library benchmark and optimal library selection,'' \emph{IEEE Transactions on Mobile Computing}, vol.~23, no.~5, pp. 5069--5082, 2024.

\bibitem{zhang2024secaas}
Z.~Zhang, C.~Ding, Y.~Li, J.~Yu, and J.~Li, ``Secaas-based partially observable defense model for iiot against advanced persistent threats,'' \emph{IEEE Transactions on Services Computing}, vol.~17, no.~6, pp. 4267--4280, 2024.

\bibitem{huang2024multi}
J.~Huang, F.~Liu, and J.~Zhang, ``Multi-dimensional qos evaluation and optimization of mobile edge computing for iot: A survey,'' \emph{Chinese Journal of Electronics}, vol.~33, no.~4, pp. 859--874, 2024.

\bibitem{yao2024intelligent}
L.~Yao, X.~Xu, W.~Dou, and M.~Bilal, ``An intelligent privacy protection scheme for efficient edge computation offloading in iov,'' \emph{Chinese Journal of Electronics}, vol.~33, no.~4, pp. 910--919, 2024.

\bibitem{Han@Deep}
S.~Han, H.~Mao, and W.~J. Dally, ``Deep compression: Compressing deep neural networks with pruning, trained quantization and huffman coding,'' in \emph{Proceedings of the International Conference on Learning Representations}, 2016, pp. 1--14.

\bibitem{Hubara@Quantized}
I.~Hubara, M.~Courbariaux, D.~Soudry, R.~EI-Yaniv, and Y.~Bengio, ``Quantized neural networks: Training neural networks with low precision weights and activations,'' \emph{Journal of Machine Learning Research}, vol.~18, no. 187, pp. 1--30, 2018.

\bibitem{banner2019post}
R.~Banner, Y.~Nahshan, and D.~Soudry, ``Post training 4-bit quantization of convolutional networks for rapid-deployment,'' in \emph{Advances in Neural Information Processing Systems}, 2019, pp. 7950--7958.

\bibitem{jin2020adabits}
Q.~Jin, L.~Yang, and Z.~Liao, ``Adabits: Neural network quantization with adaptive bit-widths,'' in \emph{Proceedings of the IEEE/CVF Conference on Computer Vision and Pattern Recognition}, 2020, pp. 2143--2153.

\bibitem{yu2021any}
H.~Yu, H.~Li, H.~Shi, T.~S. Huang, and G.~Hua, ``Any-precision deep neural networks,'' in \emph{Proceedings of the AAAI Conference on Artificial Intelligence}, vol.~35, no.~12, 2021, pp. 10\,763--10\,771.

\bibitem{xu2023eq}
K.~Xu, L.~Han, Y.~Tian, S.~Yang, and X.~Zhang, ``Eq-net: Elastic quantization neural networks,'' in \emph{Proceedings of the IEEE/CVF International Conference on Computer Vision}, 2023, pp. 1505--1514.

\bibitem{liu24spark}
F.~Liu, N.~Yang, H.~Li, Z.~Wang, Z.~Song, S.~Pei, and L.~Jiang, ``Spark: Scalable and precision-aware acceleration of neural networks via efficient encoding,'' in \emph{Proceedings of the IEEE International Symposium on High-Performance Computer Architecture}, 2024, pp. 1029--1042.

\bibitem{nagel2020up}
M.~Nagel, R.~A. Amjad, M.~Van~Baalen, C.~Louizos, and T.~Blankevoort, ``Up or down? adaptive rounding for post-training quantization,'' in \emph{Proceedings of the International Conference on Machine Learning}, 2020, pp. 7197--7206.

\bibitem{hubara2021accurate}
I.~Hubara, Y.~Nahshan, Y.~Hanani, R.~Banner, and D.~Soudry, ``Accurate post training quantization with small calibration sets,'' in \emph{Proceedings of the International Conference on Machine Learning}, 2021, pp. 4466--4475.

\bibitem{li2021brecq}
Y.~Li, R.~Gong, X.~Tan, Y.~Yang, P.~Hu, Q.~Zhang, F.~Yu, W.~Wang, and S.~Gu, ``Brecq: Pushing the limit of post-training quantization by block reconstruction,'' in \emph{Proceedings of the International Conference on Learning Representations}, 2021.

\bibitem{frantar2022optimal}
E.~Frantar, S.~P. Singh, and D.~Alistarh, ``Optimal brain compression: A framework for accurate post-training quantization and pruning,'' in \emph{Advances in Neural Information Processing Systems}, 2022, pp. 4475--4488.

\bibitem{guo2022squant}
C.~Guo, Y.~Qiu, J.~Leng, X.~Gao, C.~Zhang, Y.~Liu, F.~Yang, Y.~Zhu, and M.~Guo, ``{SQ}uant: On-the-fly data-free quantization via diagonal hessian approximation,'' in \emph{Proceedings of the International Conference on Learning Representations}, 2022.

\bibitem{Stefanos@SPINN}
S.~Laskaridis, S.~I. Venieris, M.~Almeida, I.~Leontiadis, and N.~D. Lane, ``Spinn: Synergistic progressive inference of neural networks over device and cloud,'' in \emph{Proceedings of 26th Annual International Conference on Mobile Computing and Networking}, 2020, pp. 488--502.

\bibitem{yang2022cnnpc}
S.~Yang, Z.~Zhang, C.~Zhao, X.~Song, S.~Guo, and H.~Li, ``Cnnpc: End-edge-cloud collaborative cnn inference with joint model partition and compression,'' \emph{IEEE Transactions on Parallel and Distributed Systems}, vol.~33, no.~12, pp. 4039--4056, 2022.

\bibitem{chen2023energy}
R.~Chen, L.~Li, K.~Xue, C.~Zhang, M.~Pan, and Y.~Fang, ``Energy efficient federated learning over heterogeneous mobile devices via joint design of weight quantization and wireless transmission,'' \emph{IEEE Transactions on Mobile Computing}, vol.~22, no.~12, pp. 7451--7465, 2023.

\bibitem{zhang23lightfr}
H.~Zhang, F.~Luo, J.~Wu, X.~He, and Y.~Li, ``Lightfr: Lightweight federated recommendation with privacy-preserving matrix factorization,'' \emph{ACM Transactions on Information Systems}, vol.~41, no.~4, pp. 1--28, 2023.

\bibitem{chen2024comm}
C.~Yang, J.~Yuan, Y.~Wu, Q.~Sun, A.~Zhou, S.~Wang, and M.~Xu, ``Communication-efficient satellite-ground federated learning through progressive weight quantization,'' \emph{IEEE Transactions on Mobile Computing}, vol.~23, no.~9, pp. 8999--9011, 2024.

\bibitem{fang2018nestdnn}
B.~Fang, X.~Zeng, and M.~Zhang, ``Nestdnn: Resource-aware multi-tenant on-device deep learning for continuous mobile vision,'' in \emph{Proceedings of the 24th Annual International Conference on Mobile Computing and Networking}, 2018, pp. 115--127.

\bibitem{Fang@FlexDNN}
B.~Fang, X.~Zeng, F.~Zhang, H.~Xu, and M.~Zhang, ``Flexdnn: Input-adaptive on-device deep learning for efficient mobile vision,'' in \emph{Proceedings of 5th IEEE/ACM Symposium on Edge Computing}, 2020, pp. 84--95.

\bibitem{wen2023adaptivenet}
H.~Wen, Y.~Li, Z.~Zhang, S.~Jiang, X.~Ye, Y.~Ouyang, Y.~Zhang, and Y.~Liu, ``Adaptivenet: Post-deployment neural architecture adaptation for diverse edge environments,'' in \emph{Proceedings of the 29th Annual International Conference on Mobile Computing and Networking}, 2023, pp. 408--424.

\bibitem{ding2021resource}
C.~Ding, A.~Zhou, X.~Liu, X.~Ma, and S.~Wang, ``Resource-aware feature extraction in mobile edge computing,'' \emph{IEEE Transactions on Mobile Computing}, vol.~21, no.~1, pp. 321--331, 2022.

\bibitem{ding2022resource}
C.~Ding, Y.~Li, Z.~Lu, S.~Wang, and S.~Guo, ``A resource-efficient feature extraction framework for image processing in iot devices,'' \emph{IEEE Transactions on Mobile Computing}, vol.~23, no.~1, pp. 42--55, 2024.

\bibitem{ding2022edge}
C.~Ding, Y.~Li, and S.~Wang, ``Edge/cloud-assisted feature extraction in iot devices,'' \emph{IEEE Internet of Things Journal}, vol.~9, no.~21, pp. 21\,594--21\,606, 2022.

\bibitem{machinesGFLOPS}
\BIBentryALTinterwordspacing
The gflops/w of the various machines in the vmw research group. [Online]. Available: \url{https://web.eece.maine.edu/~vweaver/group/green_machines.html}
\BIBentrySTDinterwordspacing

\bibitem{jetsonhardware}
\BIBentryALTinterwordspacing
Nvidia jetson technical specifications. [Online]. Available: \url{https://developer.nvidia.com/embedded/jetson-modules}
\BIBentrySTDinterwordspacing

\bibitem{tflite}
\BIBentryALTinterwordspacing
Tensorflow lite: Tensorflow's lightweight solution for mobile and embedded devices. [Online]. Available: \url{https://www.tensorflow.org/lite}
\BIBentrySTDinterwordspacing

\bibitem{pytorchmobile}
\BIBentryALTinterwordspacing
Pytorch mobile: End-to-end workflow from training to deployment for ios and android mobile devices. [Online]. Available: \url{https://pytorch.org/mobile}
\BIBentrySTDinterwordspacing

\bibitem{ncnn}
\BIBentryALTinterwordspacing
Ncnn: A high-performance neural network inference computing framework optimized for mobile platforms. [Online]. Available: \url{https://github.com/Tencent/ncnn}
\BIBentrySTDinterwordspacing

\bibitem{onnx}
\BIBentryALTinterwordspacing
Onnx: Open standard for machine learning interoperability. [Online]. Available: \url{https://github.com/onnx/onnx}
\BIBentrySTDinterwordspacing

\bibitem{onnxruntime}
\BIBentryALTinterwordspacing
Onnx runtime: Cross-platform, high performance ml inferencing and training accelerator. [Online]. Available: \url{https://github.com/microsoft/onnxruntime}
\BIBentrySTDinterwordspacing

\bibitem{petersen2022difflogic}
F.~Petersen, C.~Borgelt, H.~Kuehne, and O.~Deussen, ``Deep differentiable logic gate networks,'' in \emph{Advance in Neural Information Processing Systems}, 2022, pp. 2006--2018.

\bibitem{petersen2023distributional}
\BIBentryALTinterwordspacing
F.~Petersen and T.~Sutter. {Distributional Quantization}. [Online]. Available: \url{https://github.com/Felix-Petersen/distquant}
\BIBentrySTDinterwordspacing

\bibitem{deng2009imagenet}
J.~Deng, W.~Dong, R.~Socher, L.-J. Li, K.~Li, and L.~Fei-Fei, ``Imagenet: A large-scale hierarchical image database,'' in \emph{Proceedings of the IEEE Conference on Computer Vision and Pattern Recognition}, 2009, pp. 248--255.

\bibitem{He@Deep}
K.~He, X.~Zhang, S.~Ren, and J.~Sun, ``Deep residual learning for image recognition,'' in \emph{Proceedings of the IEEE/CVF Conference on Computer Vision and Pattern Recognition}, 2016, pp. 770--778.

\bibitem{huang2017densely}
G.~Huang, Z.~Liu, L.~Van Der~Maaten, and K.~Q. Weinberger, ``Densely connected convolutional networks,'' in \emph{Proceedings of the IEEE Conference on Computer Vision and Pattern Recognition}, 2017, pp. 2261--2269.

\bibitem{xie2017Aggregated}
S.~Xie, R.~Girshick, P.~Dollár, Z.~Tu, and K.~He, ``Aggregated residual transformations for deep neural networks,'' in \emph{Proceedings of the IEEE Conference on Computer Vision and Pattern Recognition}, 2017, pp. 5987--5995.

\bibitem{Howard@MobileNets}
A.~G. Howard, M.~Zhu, B.~Chen, D.~Kalenichenko, W.~Wang, T.~Weyand, M.~Andreetto, and H.~Adam, ``Mobilenets: Efficient convolutional neural networks for mobile vision applications,'' in \emph{arXiv preprint arXiv:1704.04861}, 2017.

\bibitem{Sandler@MobileNetV2}
M.~Sandler, A.~G. Howard, M.~Zhu, A.~Zhmoginov, and L.-C. Chen, ``Mobilenetv2: Inverted residuals and linear bottlenecks,'' in \emph{Proceedings of the IEEE/CVF Conference on Computer Vision and Pattern Recognition}, 2018, pp. 4510--4520.

\bibitem{Zhang@ShuffleNet}
X.~Zhang, X.~Zhou, M.~Lin, and J.~Sun, ``Shufflenet: An extremely efficient convolutional neural network for mobile devices,'' in \emph{Proceedings of the IEEE/CVF Conference on Computer Vision and Pattern Recognition}, 2018, pp. 6848--6856.

\bibitem{Ma@ShuffleNetV2}
N.~Ma, X.~Zhang, H.-T. Zheng, and J.~Sun, ``Shufflenet v2: Practical guidelines for efficient cnn architecture design,'' in \emph{Proceedings of the European Conference on Computer Vision}, 2018, pp. 116--131.

\bibitem{tan2019efficientnet}
M.~Tan and Q.~Le, ``{E}fficient{N}et: Rethinking model scaling for convolutional neural networks,'' in \emph{Proceedings of the International Conference on Machine Learning}, 2019, pp. 6105--6114.

\bibitem{Dosovitskiy@vit}
A.~Dosovitskiy, L.~Beyer, A.~Kolesnikov, D.~Weissenborn, X.~Zhai, T.~Unterthiner, M.~Dehghani, M.~Minderer, G.~Heigold, S.~Gelly, J.~Uszkoreit, and N.~Houlsby, ``An image is worth 16x16 words: Transformers for image recognition at scale,'' in \emph{Proceedings of the International Conference on Learning Representations}, 2021, pp. 1--21.

\bibitem{touvron21deit}
H.~Touvron, M.~Cord, M.~Douze, F.~Massa, A.~Sablayrolles, and H.~Jegou, ``Training data-efficient image transformers \& distillation through attention,'' in \emph{Proceedings of the International Conference on Machine Learning}, 2021, pp. 10\,347--10\,357.

\bibitem{liu2021swin}
Z.~Liu, Y.~Lin, Y.~Cao, H.~Hu, Y.~Wei, Z.~Zhang, S.~Lin, and B.~Guo, ``Swin transformer: Hierarchical vision transformer using shifted windows,'' in \emph{Proceedings of the IEEE/CVF International Conference on Computer Vision}, 2021, pp. 9992--10\,002.

\end{thebibliography}

\end{document}